\documentclass[journal]{IEEEtran}
\ifCLASSINFOpdf
\else
\fi

\usepackage{url}
\usepackage{graphicx}
\usepackage{algorithmic}
\usepackage{subfigure}
\usepackage{amsmath}
\usepackage{bbm}
\usepackage{xcolor}
\usepackage{makecell}
\usepackage{booktabs}
\usepackage{multirow}
\usepackage{balance}
\usepackage{listings}
\usepackage{booktabs}
\usepackage{pifont}
\usepackage{enumitem}
\usepackage[ruled,vlined,linesnumbered]{algorithm2e}
\usepackage{bm}
\usepackage{array}
\usepackage{ragged2e}
\usepackage{tabularray}
\usepackage[normalem]{ulem}
\usepackage{enumitem}
\usepackage{pythonhighlight}
\usepackage[most]{tcolorbox}
\usepackage[table]{xcolor}
\usepackage{soul}

\sethlcolor{lightgray}
\newtcbtheorem[]{cmt}{P.}{
	colbacktitle = black!60!white, colframe = black!60!white,
	colback = black!3!white,
fonttitle=\bfseries\small,
        fontupper = \small,
        right=3pt, left=3pt, top=1pt, bottom=1pt
}{t}

\newtcolorbox{myquote}{
  colback=gray!10,
  colframe=gray!80,
  left=0.6em,
  right=0.6em,
  top=0.3em,
  bottom=0.3em,
  fontupper=\fontsize{8}{9}\selectfont
}

\newcommand{\ie}{\textit{i}.\textit{e}.}
\newcommand{\eg}{\textit{e}.\textit{g}.}

\newcommand{\project}{\texttt{SE\textsuperscript{2}}\xspace}
\definecolor{light_blue}{HTML}{f1ffff}

\begin{document}

\title{A Self-Exploring and Self-Evolving Paradigm for Adaptive Edge AI Development}

\author{Leming Shen,
Qiang Yang,
Xinyu Huang,
Zijing Ma, 
Yuanqing Zheng, 
Chris Xiaoxuan Lu
\thanks{L. Shen, X. Huang, Z. Ma, and Y. Zheng are with the Department of Computing, The Hong Kong Polytechnic University, Hong Kong SAR, China.
E-mail: leming.shen@connect.polyu.hk, unixy-xinyu.huang@connect.polyu.hk, zijing.ma@connect.polyu.hk, and csyqzheng@comp.polyu.edu.hk.}%
\thanks{Q. Yang is with the Department of Computer Science and Technology,
University of Cambridge, Cambridge, UK.
E-mail: qiang.yang@cl.cam.ac.uk.}
\thanks{C. Lu is with the Department of Computer Science, University College London, UK.
E-mail: xiaoxuan.lu@ucl.ac.uk.}
}

\markboth%
{IEEE TRANSACTIONS ON MOBILE COMPUTING,~Vol.~X, No.~X, September~2026}%
{Shen \MakeLowercase{\textit{et al.}}: \project: Automating Model Deployment on Heterogeneous Embedded Devices}

\maketitle

\begin{abstract}
Large Language Models (LLMs) exhibit powerful code generation abilities, fostering agentic systems to automate edge AI application development. However, existing agents typically operate via fixed agentic workflows and predefined design spaces, struggling to handle rapidly evolving demands of real-world scenarios (\eg, heterogeneous data modalities, diverse task pipelines). Unlike prior systems that treat LLMs mainly as \textit{code generators} following rigid workflows, we fully exploit their \textit{reasoning capabilities} to allow agents to determine their own workflows on the fly. To this end, we propose \project, a self-exploring and self-evolving paradigm for adaptive edge AI development. Inspired by human learning processes, we distill development into four atomic operations (\texttt{retrieve}, \texttt{reason}, \texttt{code}, \texttt{test}), and empower agents to explore workflows by dynamically and adaptively selecting an appropriate next operation based on the current context. Extensive evaluations show that, with adaptive workflows, the synthesized programs achieve up to 26\% higher accuracy and consume 28\% lower resources than state-of-the-art baselines, while maintaining \textit{superior adaptability} to unforeseen edge AI scenarios. This highlights the potential of reasoning-driven adaptive agentic workflow in edge AI development.

\end{abstract}

\begin{IEEEkeywords}
Autonomous coding, LLM, agent
\end{IEEEkeywords}

\maketitle

\section{Introduction}
\label{sec:introduction}

Edge AI integrates AI techniques with heterogeneous mobile devices to enable perception and actuation in real-world environments \cite{siam2025artificial, wan2025meda, gong2025data, cheng2025integrated, lee2025hypercam, DBLP:journals/tmc/LiuWLCZS24}, facilitating various applications such as smart sensing \cite{cui2024towards}, healthcare \cite{shao2025my, shao2024joey}, and autonomous driving \cite{hu2024t, do2025lightweight, ahmad2022idriving, cui2025towards}. 

Developing such applications, however, relies on substantial domain expertise among developers, who must analyze multimodal datasets, design complex signal-processing pipelines, train AI models under diverse constraints, and deploy them on heterogeneous mobile devices \cite{DBLP:journals/tmc/ZhangYZ24, DBLP:journals/tmc/WangZL26, DBLP:journals/tmc/CortiMSPS26, DBLP:journals/tmc/FangGLZTMLSY26, DBLP:journals/tmc/YiZYWLH26, 11603432, shen2024fedconv}. Therefore, recent works \cite{shen2025autoiot, yang2026autoembed, 10.1145/3636534.3697447} build LLM-based agentic systems to automate this process by leveraging LLMs' \textit{code generation capabilities} and embedded domain expertise.
These systems can translate user requirements into executable edge AI programs based on handcrafted, predefined agentic workflows, chaining modules such as knowledge retrieval, task planning, code generation, and iterative refinement \cite{shang2025agentsquare, hu2025automated}. 

\begin{figure}
    \centering
    \includegraphics[width=0.48\textwidth]{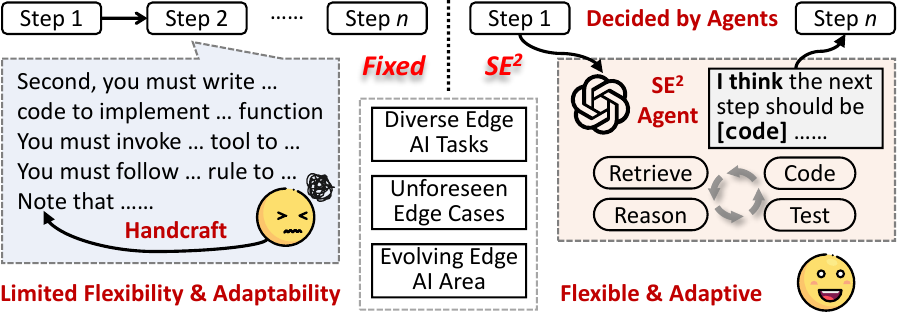}
    \caption{Previous works handcrafted fixed agentic workflows to synthesize programs. \project autonomously explores and selects the next atomic operation in program synthesis to meet diverse edge AI requirements.}
    \label{fig:agent_human_comparison}
\end{figure}

Fixed agentic workflows, however, often struggle to handle diverse edge AI tasks in real-world scenarios (Fig.~\ref{fig:agent_human_comparison}, left). In practice, edge AI tasks may span numerous mobile devices, heterogeneous runtime environments, and dynamic performance and resource constraints. As our study reveals (\S~\ref{sec:preliminary_study}), even well-engineered agentic systems may struggle to accommodate such variability, suffering from \textit{inflexibility}, \textit{limited adaptability}, and \textit{high maintenance costs}.

The root cause is that fixed workflows are often \textit{constrained by the specific assumptions and development experiences of workflow designers}. In practice, developers must explicitly enumerate workflow steps and handcraft detailed prompts that regulate agent actions for each. As a result, LLMs solely function as \textit{code generators} following predefined instruction flows. Yet, a key strength of LLMs remains under-explored: their \textit{reasoning capabilities}, \ie, the ability to make decisions in response to dynamic context information. This gap motivates a key research question: \textit{Can we harness LLMs' reasoning capabilities to autonomously construct adaptive workflows to meet the ever-changing requirements of edge AI development?} 
If successful, we could shift the burden of workflow design from developers to AI agents, enabling more flexible, adaptive, and scalable edge AI application development.

Moving toward such a \underline{s}elf-\underline{e}xploring and \underline{s}elf-\underline{e}volving agentic paradigm, we propose \project that leverages LLMs' reasoning abilities to autonomously construct distinct workflows for various edge AI tasks. The core insight is inspired by human learning processes, which involve iterative knowledge acquisition, reasoning, practice, and reflection on accumulated experiences. Likewise, an \project agent explores its workflow and evolves through iterations of reflection and exploration. Building upon this vision, we distill edge AI development into four atomic operations (\texttt{retrieve}, \texttt{reason}, \texttt{code}, \texttt{test}), which serve as meta building blocks of agentic workflows (Fig.~\ref{fig:agent_human_comparison}, right). The \project agent autonomously selects the next appropriate atomic operation based on its reasoning about the accumulated context, thereby flexibly adapting to diverse edge AI scenarios. In addition, our \project agent evolves through self-reflection, rather than blindly adhering to rigid workflows.

Nevertheless, three unique technical challenges needed to be addressed, calling for novel design and implementation:

\noindent\ding{182} \textbf{Limited long-context reasoning capability.}
It is inefficient and impractical to rely on a single agent to perform all the atomic operations due to the limited long-context reasoning abilities of current LLMs, even if they have large context windows \cite{moon2025needlechain}. 
To tackle this, we adopt a \textit{coordinator-actor agentic architecture}, where a \textit{coordinator} agent maintains a global view and autonomously orchestrates four task-specific atomic agents, \ie, \textit{retriever}, \textit{reasoner}, \textit{programmer}, and \textit{tester}. 
As detailed in \S~\ref{sec:detailed_design}, this modular division of cognitive tasks effectively reduces reasoning complexity over long contexts, fostering interpretable and traceable decision-making.

\noindent\ding{183} \textbf{Lack of in-depth understanding of edge devices.}
Our feasibility study (\S~\ref{sec:feasibility_study}) found that, even augmented with detailed device specifications, LLMs still struggle to generate executable and performant edge AI programs. The root cause is that LLMs lack runtime feedback from real-world mobile devices. Consequently, the generated code often overlooks critical operational semantics and hardware constraints (\eg, CPU architectures, environment affinity, and memory allocation). To address this issue, we propose a \textit{mobile runtime feedback mechanism} that encodes grounded, runtime-level semantics and conveys them to \project agents as actionable feedback, thereby closing the loop for optimization and iteration.

\noindent\ding{184} \textbf{Inability to adapt to evolving Edge AI tasks.}
Simply relying on vanilla LLMs for decision-making cannot cope with the emerging and evolving edge AI environments and tasks.
To enhance the adaptability of the \textit{coordinator} as well as the atomic agents, we adopt reinforcement fine-tuning (RFT) \cite{RFT} with an \textit{ensemble mutual-actor-critic strategy} to improve their domain-specific reasoning abilities. 
In addition, we design several continuously self-evolving graders to assess the reasoning quality in edge AI contexts and provide adaptive feedback to the RFT loop.
This approach enables agents to acquire new domain knowledge over time and agilely adjust their workflows to handle unseen scenarios and edge cases.

We fully implement \project and comprehensively evaluate it against 8 SOTA baselines on an edge AI code-generation benchmark and three representative \textit{multimodal} edge AI applications (\textit{human activity recognition}, \textit{vision-language understanding}, and \textit{autonomous driving}), spanning distinct modalities, reasoning complexity, and real-world constraints. 
Extensive experiments reveal that baseline coding agents cannot consistently generate performant code (40\% success rate lower than \project). 
In contrast, \project achieves up to 26\% higher accuracy and 28\% reduction in resource consumption compared to SOTA methods, while retaining superior \textit{generalizability} across tasks. These results validate the effectiveness of our \textit{reasoning-driven} adaptive workflow construction paradigm.

In summary, we make the following contributions:

\begin{figure}[t]
    \centering
    \includegraphics[width=0.45\textwidth]{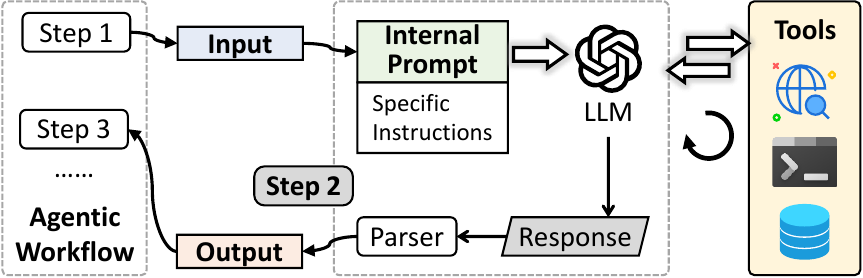}
    \caption{A typical agentic workflow of existing work with detailed step 2.}
    \label{fig:agentic_workflow_background}
\end{figure}

\begin{itemize}[leftmargin=9pt]
    \item \project exploits LLMs' reasoning capabilities to autonomously construct agentic workflows via self-exploration and self-evolution, addressing the limitations of fixed workflows in heterogeneous and evolving edge AI environments.
    \item We design a reasoning enhancement framework featuring atomic-operation-level adaptive agentic workflow construction, runtime-grounded probe synthesis, and continual multi-agent reasoning evolution. 
    \item We implement and extensively evaluate \project against eight SOTA agents on a public benchmark, three representative multimodal edge AI applications, and through a user study, demonstrating its superior accuracy and generalizability.
\end{itemize}

\section{Background \& Motivation}
\label{sec:background}

\subsection{Limitations of SOTA Methods}
\label{sec:preliminary_study}

By equipping LLMs with external tools (\eg, search engines), we can build agents \cite{wen2025autodroid, lee2024mobilegpt, nguyen2025graspmas, wang2025cmp, zhang2025position, lee2025llm, dong2024creating, choi2025collective, choi2025private, zhang2025towards} and design agentic workflows guiding them to automatically solve complex tasks.
As shown in Fig.~\ref{fig:agentic_workflow_background}, an agentic workflow is typically a multi-step process where each step receives the output from the previous step and produces a new intermediate result. 
With carefully crafted prompts and tool usage rules in each step, the agent can follow the workflow and synthesize various executable edge AI programs for users.

To better understand the limitations (\ie, \textit{inflexibility}, \textit{limited adaptability}, and \textit{high maintenance costs}) of existing agentic workflows, we conduct a preliminary study by reproducing three SOTA works for edge AI development: AutoIOT \cite{shen2025autoiot}, AutoEmbed \cite{yang2026autoembed}, and IoTPilot \cite{gong2025programming}. These methods adopt predefined workflows involving background knowledge retrieval, task decomposition, code generation, and debugging. 

\begin{figure}[t]
    \centering
    \subfigure[Existing agentic systems typically fail to handle unforeseen scenarios]{
        \includegraphics[width=0.225\textwidth]{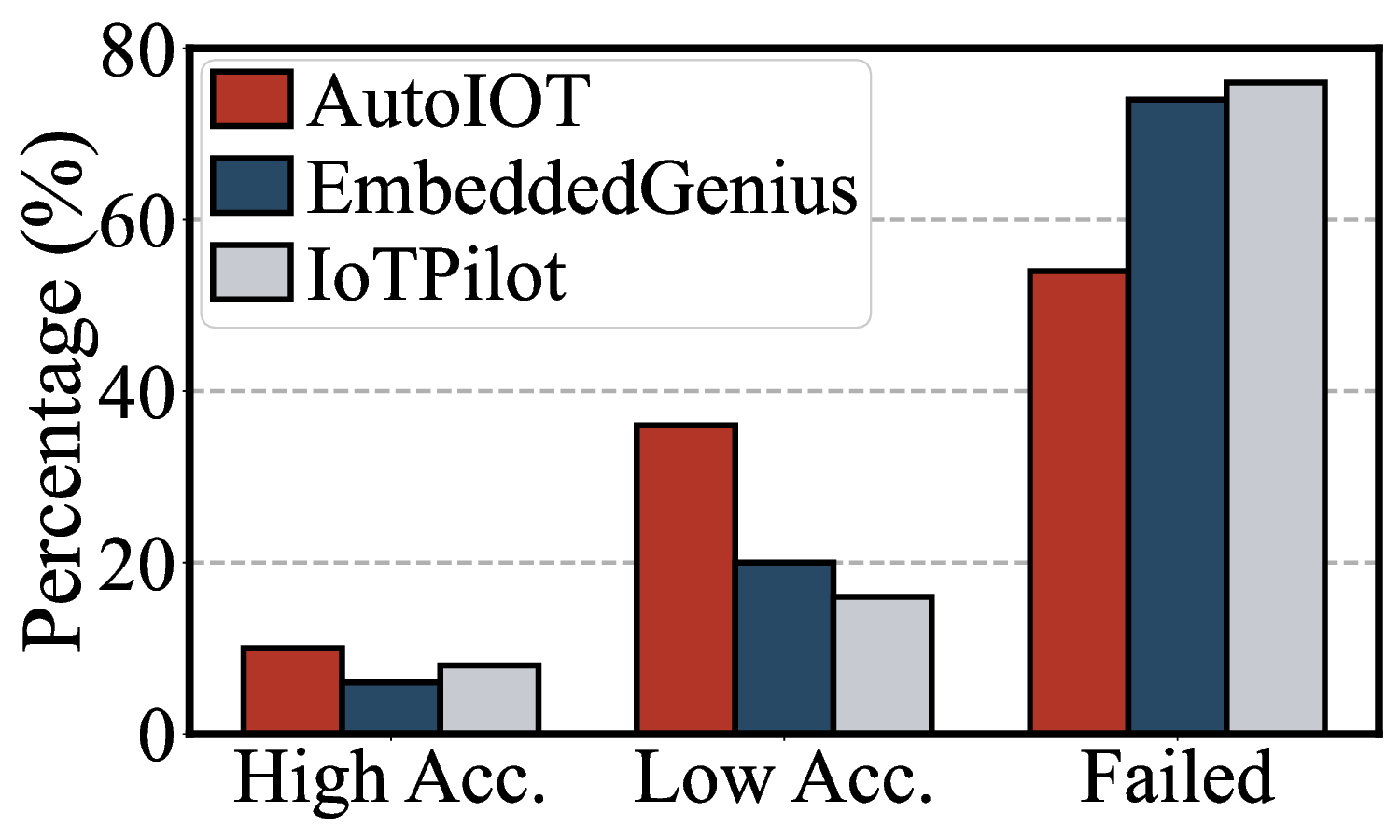}
        \label{fig:preliminary_study}
    }
    \centering
    \subfigure[Though feasible, most agents exhibited limited reasoning capabilities]{
        \includegraphics[width=0.225\textwidth]{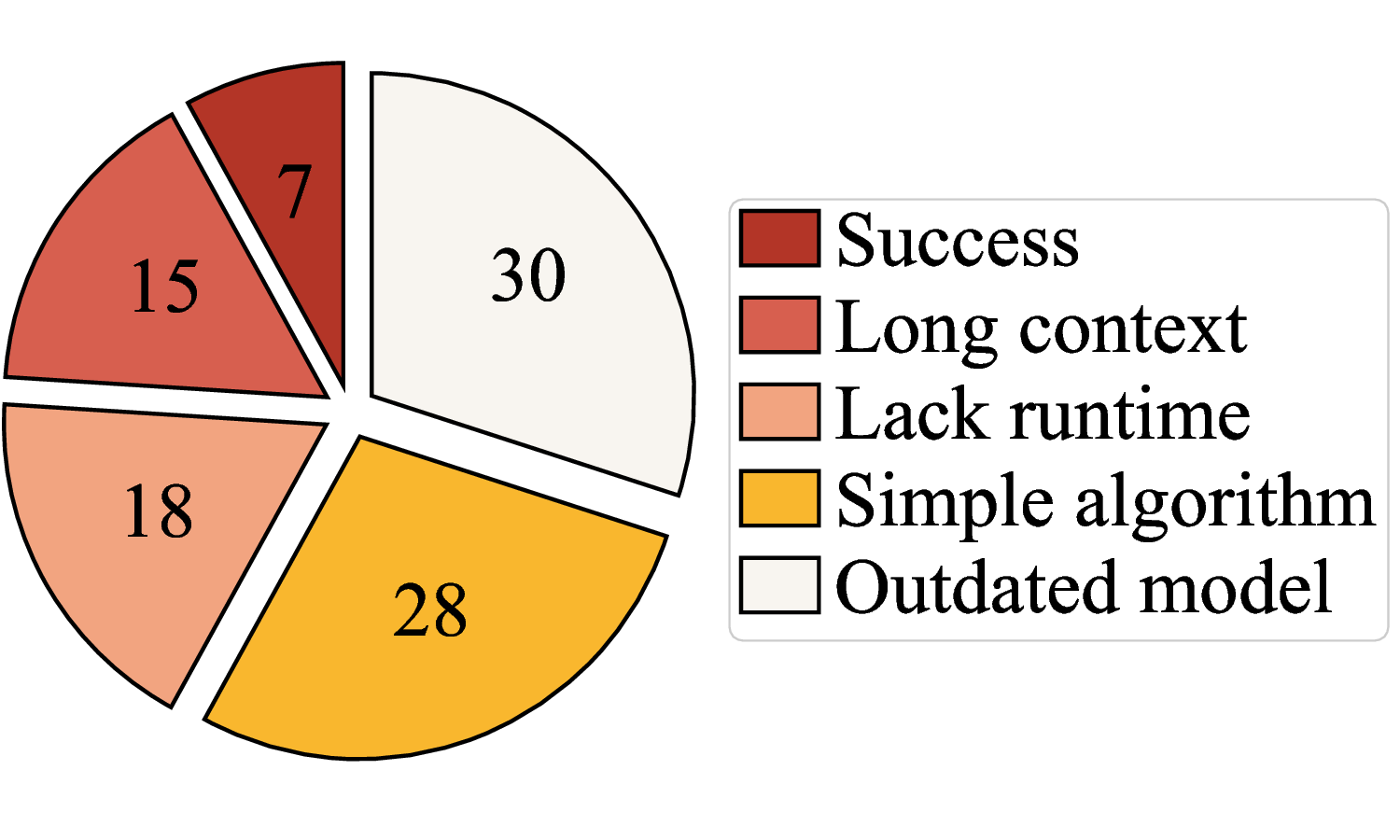}
        \label{fig:feasibility_study}
    }
    \caption{Preliminary \& feasibility study.}
\end{figure}

We ask them to develop a human activity recognition system (HAR) using the multimodal XRF55 dataset \cite{wang2024xrf55} and repeat 100 times to avoid randomness. We randomly remove data to simulate the modality missing issue in real-world multimodal scenarios \cite{zheng2023autofed}. As shown in Fig.~\ref{fig:preliminary_study}, \textbf{only 32\% of the synthesized programs of SOTA methods can be successfully executed}, and only 8\% achieve comparable performance to a SOTA handcrafted program. 
\textbf{It is highly challenging to design a fixed, one-size-fits-all workflow to adapt to unforeseen cases and accommodate diverse edge AI requirements}. 
Further inspection reveals heterogeneous issues:

\begin{itemize}[leftmargin=9pt]
    \item \textit{No unimodal verification}: They do not verify the correctness of each code snippet for unimodal data processing before multimodal fusion. This is often discouraged in current multimodal application development pipelines \cite{ouyang2023harmony}.
    \item \textit{No handling of missing modalities}: 68\% of the programs fail when certain data streams are absent due to the lack of elastic AI models for dynamic input handling \cite{zheng2023autofed}.
\end{itemize}

\noindent To remedy the above as well as other heterogeneous failures, developers have to repeatedly revise and refine their workflows by explicitly listing regulations 
such as tool usage rules, edge-case handling advice, and consistency checks. Maintaining such crafted rules for unseen scenarios with diverse requirements becomes highly time-consuming and error-prone.


\subsection{Motivation \& Feasibility Study}
\label{sec:feasibility_study}

\noindent\textbf{Motivation}.
The root cause of these limitations is that the workflow design and its underlying reasoning still heavily rely on human developers, while LLMs are merely treated as \textit{code generators} following predefined instructions. 
As a result, these fixed workflows cannot adapt to new tasks in practical real-world scenarios. 
Unlike SOTA methods, \project exploits LLMs' reasoning abilities to enable agents to autonomously optimize and adjust their own workflows in response to diverse goals.

We analyze over 30 agent-based code generation systems and find that nearly all edge AI development workflows are composed of four atomic agentic operations:

\begin{itemize}[leftmargin=9pt]
    \item \texttt{\textbf{retrieve}}: An agent retrieves relevant domain knowledge from various data sources (\eg, databases, websites).
    \item \texttt{\textbf{reason}}: An agent analyzes the current and past state based on the available information and makes inferences before selecting the next operation. This operation relies on the agent's reasoning capabilities for decision-making.
    \item \texttt{\textbf{code}}: An agent generates a Python\footnote{We focus on Python since it is widely used for AI development \cite{prabu2024contribution}.} code snippet.
    \item \texttt{\textbf{test}}: An agent generates test cases, deploys the code snippet on the target mobile device (optional), executes and evaluates it, and records the evaluation results.
\end{itemize}
\textbf{By enabling AI agents to autonomously select among these operations in response to changing states, agentic workflows can be organically constructed.}
Similar to next token prediction in LLMs, \project instead predicts the next atomic operation that drives the workflow forward.

\noindent\textbf{Feasibility \& Challenges}.
We build a single agent to test whether it can construct workflows autonomously.
We instruct the agent to synthesize a multimodal HAR program and deploy it on a Jetson Nano board.
We repeat the test 100 times and find that certain constructed workflows can sometimes generate executable edge AI code with satisfactory performance after multiple trials. Notably, they share a similar and commonly adopted structure: "retrieve knowledge $\rightarrow$ generate code snippets $\rightarrow$ test individually $\rightarrow$ multimodal fusion $\rightarrow$ build elastic AI models $\rightarrow$ iteratively refine". This implies that adaptive workflow is feasible.
Despite this feasibility, the overall success rate remains low (8/100), comparable to directly running existing agents without modification. Further in-depth analysis reveals that the failures stem from three fundamental reasoning limitations (Fig.~\ref{fig:feasibility_study}), precisely motivating the technical challenges that \project need to overcome:

\begin{itemize}[leftmargin=8pt]\setlength{\itemsep}{2pt} 
    \item \textbf{Long-Context Reasoning.} 16 agents repeatedly execute the same action (\eg, searching dataset information) and fail to terminate. It is likely because a single agent performs all operations and accumulates increasingly long contexts, while LLMs inherently pay more attention to the earliest and most recent tokens in the context window, causing the agent to "forget" previously executed steps \cite{zhang2024efficient}.
    \item \textbf{Runtime-Aware Reasoning.} Even with retrieved knowledge of the target device, 18 agents still produce error code either because the required runtime environments are incompatible with the device or because the resource consumption exceeds the device's capacity. The reason is that the agent cannot access the device and thus lacks an in-depth understanding of its runtime environments \cite{DBLP:journals/tmc/VenierisALL24} and resource constraints.
    \item \textbf{Outdated Models.} 58 agents adopt overly general or outdated algorithms and AI models. This indicates that LLMs lack the reasoning capabilities to associate user requirements with SOTA edge AI algorithms and models.
\end{itemize}
To address these challenges, we design and implement a series of novel techniques as detailed in \S~\ref{sec:detailed_design}.

\section{\project Overview}
\label{sec:problem_formulation}

\begin{figure*}[t]
    \centering
    \includegraphics[width=0.87\textwidth]{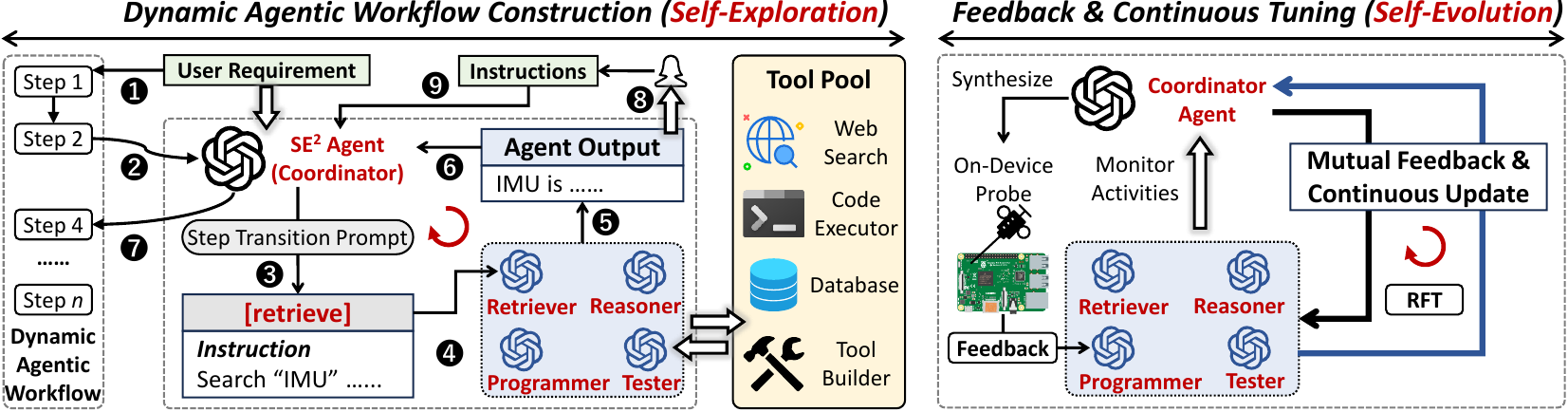}
    \caption{The overall architecture and workflow of \project (with detailed operations in step 3).}
    \label{fig:system_overview}
\end{figure*}

\project aims to enhance and exploit LLMs' reasoning abilities to dynamically and adaptively construct agentic workflows, which translate user requests into executable edge AI code on target devices. 
As shown in Fig.~\ref{fig:system_overview}, \project has two stages: 1) \textit{self-exploration} for dynamic workflow construction, and 2) \textit{self-evolution} for runtime feedback and continuous LLM tuning.

\noindent\textbf{Self-Exploration.} 
Given a user-requested edge AI task (\ding{182}), the \textit{coordinator} iteratively determines the next appropriate operation. At each step, the \textit{coordinator} analyzes the output from the previous operation (\ding{183}), and generates the next atomic operation with detailed instructions (\ding{184}). The corresponding atomic agent is activated (\ding{185}) to execute the operation based on the instruction and generate an output (\ding{186}), which is further sent to the \textit{coordinator} (\ding{187}). Based on the currently constructed workflow, the \textit{coordinator} decides if extra operations are required (\ding{188}) or concludes that the synthesized edge AI program meets the user requirement with acceptable performance (\ding{189}). In addition, \project provides an interface for users to offer instructions for workflow refinement (\ding{190}). 
As such, agentic workflows are autonomously determined with minimal user intervention, offering enhanced flexibility and adaptability. Note that to better interpret user intents, we can orthogonally integrate existing arts (\eg, LLM-as-a-Judge \cite{zheng2023judging}). 

\noindent\textbf{Self-Evolution.}
When the target device runs the generated code, the \textit{mobile runtime feedback mechanism} synthesizes a probe to extract runtime semantics (\eg, memory allocation) as grounded, runtime-level feedback.
In parallel, our \textit{continuous LLM tuning paradigm} records the mutual feedback between the \textit{coordinator} and atomic agents to incrementally enhance their reasoning capabilities in the edge AI domain. As such, all agents can continuously evolve, enabling them to adapt to unforeseen scenarios and emerging edge AI knowledge.

\section{Detailed Design of \project}
\label{sec:detailed_design}

\subsection{Hierarchical Coordinator-Actor Agentic Architecture}
\label{sec:master_slave_agentic_architecture}
This module performs \textbf{self-exploration} by dynamically constructing adaptive workflows. 
Due to LLMs' limited long-context reasoning abilities (\S~\ref{sec:feasibility_study}), we propose a \textit{coordinator-actor agentic architecture}, where a \textit{coordinator} agent orchestrates the self-exploring process by delegating distinct tasks to four atomic agents: \textit{retriever}, \textit{reasoner}, \textit{programmer}, and \textit{tester}. This design offloads the execution of atomic operations to multiple task-specific agents for more efficient handling. 

\noindent\textbf{Challenge.} 
In multi-agent systems, sharing contexts among agents often leads to prohibitive costs \cite{han2024llm}, introducing a new challenge: \textit{how to enable effective and efficient communication between the \textit{coordinator} and atomic agents while ensuring context coherence?} To address this, we design tailored prompts to regulate each agent's behavior and outputs.

\begin{cmt}{\textit{Coordinator} Prompt (step transition prompt, simplified)}{}
\textbf{\ding{182} Role \& Objective}: You are a \textbf{coordinator} agent operating in a multi-agent system for edge AI development. Given a \hl{user requirement}, you need to orchestrate another four atomic agents (\textbf{retriever}, \textbf{reasoner}, \textbf{programmer}, and \textbf{tester}) to synthesize an executable program.\\
\vspace{-8pt}
\raisebox{1.8ex}{\rule{0.7\textwidth}{0.5pt}}\\
\textbf{\ding{183} Respond Rules}: You \textbf{MUST} select one of the following operations:\\
1. \textbf{Retrieve} -- activate the \textbf{retriever} to retrieve relevant knowledge\\
2. \textbf{Reason} -- activate the \textbf{reasoner} to analyze and reflect on ...\\
3. \textbf{Code} -- activate the \textbf{programmer} to write a Python script ...\\
4. \textbf{Test} -- activate the \textbf{tester} to execute and evaluate the code ...\\
5. \textbf{Quit} -- end the task if the code satisfies the user requirement.\\
\vspace{-8pt}
\raisebox{1.8ex}{\rule{0.7\textwidth}{0.5pt}}\\
\textbf{\ding{184} Current State}: 
\\The summarized activity of \hl{atomic agent}: \hl{\{......\}}\\
\vspace{-8pt}
\raisebox{1.8ex}{\rule{0.7\textwidth}{0.5pt}}\\
\textbf{\ding{185} Output Specifications}: You \textbf{MUST} generate \textbf{ONLY THREE} items:\\
1. \textbf{Operation Name} -- the name of the atomic operation for execution.\\
2. \textbf{Instructions} -- a brief instruction for the corresponding agent.\\
3. \textbf{Context} -- detailed context including previously generated ...\\
\vspace{-8pt}
\raisebox{1.8ex}{\rule{0.7\textwidth}{0.5pt}}\\
\textbf{\ding{186} Simple Example Response}\\
\text{[}retrieve\text{]}\quad Retrieve detailed information of the \hl{\{dataset name\}} dataset.\\
I need to understand the dataset first, including dimensionality, ......
\end{cmt}

\noindent\textbf{\textit{Coordinator} Prompt} (\hl{P.1}) for step transition includes: 

\begin{itemize}[leftmargin=9pt]
    \item \textit{Role \& objective.} We assign the agent a \textit{coordinator} role and provide a task description that guides it to proactively orchestrate multiple agents for edge AI development.
    \item \textit{Respond rules} include descriptions of the four atomic operations and their selection rules. We task the \textit{coordinator} with iteratively selecting an appropriate operation and activating the corresponding atomic agent for execution. We also introduce a \texttt{quit} operation that ends self-exploration if the generated code meets the user requirement. Importantly, beyond sticking to the general cycle of "\textit{retrieve} $\rightarrow$ \textit{reason} $\rightarrow$ \textit{code} $\rightarrow$ \textit{test}", we encourage the \textit{coordinator} to repeat any stage as needed, \eg, "\textit{retrieve} $\rightarrow$ \textit{reason} $\rightarrow$ \textit{retrieve}".
    \item \textit{Current state.} Atomic agents' activities are recorded during operation execution, serving as the current state and thereby enabling the \textit{coordinator} to determine the next operation with enhanced contextual coherence.
    \item \textit{Output specifications.} To ensure that atomic agents can accurately interpret the operation, we stipulate that the \textit{coordinator} only outputs three items: the operation name, a brief instruction, and relevant context information.
    \item \textit{One-shot example.} We provide a sample output to reduce ambiguity in task interpretation.
\end{itemize}

\begin{cmt}{Atomic Agent Prompt (simplified) (Example: \textit{retriever})}{}
\textbf{\ding{182} Role \& Objective}: You are a \textbf{retriever} agent operating in \hl{...} Based on the provided information, you need to retrieve relevant information from external data sources (\eg, websites).\\
\vspace{-8pt}
\raisebox{1.8ex}{\rule{0.7\textwidth}{0.5pt}}\\
\textbf{\ding{183} Operation Execution Rules}: You \textbf{MUST} follow these rules:\\
1. You \textbf{MUST} invoke multiple tools to complete the instruction ...\\
2. You \textbf{MUST} retrieve as much information as possible before ...\\
3. You \textbf{MUST} summarize the retrieved content and store it ...\\
\vspace{-8pt}
\raisebox{1.8ex}{\rule{0.7\textwidth}{0.5pt}}\\
\textbf{\ding{184} Output Specifications}: Your \textbf{MUST} output one of the types below:\\
1. \textit{function\_call} -- you need to invoke tools to gather information or ...\\
2. \textit{message} -- directly output "\text{[}finish\_retrieval\text{]}" if you believe that ...\\
\vspace{-8pt}
\raisebox{1.8ex}{\rule{0.7\textwidth}{0.5pt}}\\
\textbf{\ding{185} Instruction \& Context Information}:\\
\hl{\{coordinator\_instruction\}} \quad \hl{\{context\_information\}}
\end{cmt}

\noindent\textbf{Atomic Agent Prompt} (\hl{P.2}) comprises four parts: 

\begin{itemize}[leftmargin=9pt]
    \item \textit{Role \& objective.} We also assign distinct roles to the atomic agents, along with descriptions of their jobs.
    \item \textit{Operation execution rules.} We encourage each atomic agent to proactively and repeatedly invoke relevant tools to obtain sufficient information or conduct thorough tests.
    \item \textit{Output specifications.} We stipulate that each atomic agent can only respond with a \texttt{function\_call} with parameters to invoke a tool, or a \texttt{message} to quit the current operation.
    \item \textit{Instruction \& context information} from the \textit{coordinator}.
\end{itemize}

\noindent We instruct atomic agents to invoke tools until all necessary information is gathered. Once a "finished" \texttt{message} is generated, the atomic agent's activity is summarized and returned to the \textit{coordinator} for further decision-making.

\noindent\textbf{Remarks.} 
Through our \textit{coordinator-actor agentic architecture} with crafted prompts, the \textit{coordinator} can autonomously explore appropriate atomic operations based on its reasoning abilities. As such, agentic workflows can be dynamically constructed to cope with unforeseen cases (\eg, modality missing), exhibiting enhanced flexibility and adaptability.

\subsection{Mobile Runtime Feedback Mechanism}
\label{sec:agentic_checkpointing_mechanism}

This module facilitates \textbf{self-evolution} by providing runtime-level feedback for decision adjustment. As detailed in \S~\ref{sec:feasibility_study}, even with retrieved device information, some generated code remains unexecutable, as \project cannot directly access devices and lacks detailed runtime signals during coding and testing. Thus, we propose a \textit{mobile runtime feedback mechanism} that obtains grounded, runtime-level semantics as timely feedback.

\noindent\textbf{Challenge.}
An intuitive way to extract runtime semantics is to apply existing device profiling techniques (\eg, nn-Meter \cite{zhang2021nn}). However, they rely on rule-based scripts for runtime monitoring, which are often insufficient for diverse and rapidly evolving edge AI tasks. Instead, \project synthesizes \textit{probe programs} that execute on the target device to extract only relevant runtime semantics. Our insight is that the \textit{coordinator}, overseeing the entire development process, can serve as a supervisor for probe synthesis. The runtime semantics collected by the probe are then fed back to atomic agents for response adjustments. Specifically, we design: \textit{probe synthesis}, \textit{semantic alignment grading}, and \textit{response adjustment}.

\noindent\textbf{Probe Synthesis.}
Different from general code generation, the purpose of a probe is not to solve the coding task, but to extract latent runtime semantics that agents cannot infer from textual device specifications. Specifically, we prompt (\hl{P.3}) the \textit{coordinator} to synthesize a probe program with two main considerations before on-device code execution:

\begin{itemize}[leftmargin=9pt]
    \item \textit{Runtime semantics}.
    \textit{Environment affinity} reveals incompatibilities between library versions and hardware kernels. \textit{Memory dynamics} reflect feasibility under runtime allocation and fragmentation patterns. \textit{Execution behavior} exposes performance bottlenecks arising from scheduling, warm-up, and accelerator utilization. \textit{I/O feasibility} verifies if essential data formats are supported.
    \item \textit{Probe program requirements}. \textit{Minimality} ensures that each probe focuses on a single runtime factor, reducing confounding effects and enabling precise interpretation of observed behavior. \textit{Orthogonality} prevents probes from blending unrelated concerns, preserving the diagnostic clarity for targeted adjustments. \textit{Executability} guarantees that each synthesized probe can reliably run within the device's native software stack and provide grounded evidence. \textit{Observability} mandates that probes emit structured outputs, enabling the system to consistently extract useful runtime semantics.
\end{itemize}

\begin{cmt}{Probe  Synthesis Prompt (simplified)}{}
\textbf{\ding{182} Role \& Objective}: You are a \textbf{probe-synthesizer}. Your goal is to develop a probe program that extracts relevant runtime semantics ...\\
\vspace{-8pt}
\raisebox{1.8ex}{\rule{0.7\textwidth}{0.5pt}}\\
\textbf{\ding{183} Runtime Semantics}. (tailored to the \textit{coordinator}'s instruction):\\
1. \textit{Environment affinity}; (\eg, PyTorch versions, CUDA/cuDNN bindings)\\
2. \textit{Memory dynamics} (\eg, allocation failure, fragmentation, peak usage)\\
3. \textit{Execution behavior} (\eg, latency, warm-up stability)\\
4. \textit{I/O feasibility} (\eg, data loading, sensor interface availability)\\
\vspace{-8pt}
\raisebox{1.8ex}{\rule{0.7\textwidth}{0.5pt}}\\
\textbf{\ding{184} Probe Program Requirements}:\\
1. \textit{Minimality};\quad 2. \textit{Orthogonality};\quad 3. \textit{Executability};\quad 4. \textit{Observability}\\
\vspace{-8pt}
\raisebox{1.8ex}{\rule{0.7\textwidth}{0.5pt}}\\
\textbf{\ding{185} Instruction \& Context Information}:\\
\hl{\{coordinator\_instruction\}} \quad \hl{\{context\_information\}}
\end{cmt}

\noindent\textbf{Semantic Alignment Grading}.
Based on the runtime semantics obtained from the probe, we need to judge if each atomic agent's decision aligns with the current device runtime and the edge AI requirement. Such semantic alignment manifests in two aspects: 1) \textit{Local alignment}: whether the atomic agent's response contributes to the agent's current task, \ie, the \textit{coordinator}'s instruction; 2) \textit{Global alignment}: whether it contributes to the \textit{coordinator}'s overarching task, \ie, the user requirement. We prompt the \textit{coordinator} to assign two scores reflecting the local and global alignment of each atomic agent's response based on six designed grading levels. In addition, we instruct the \textit{coordinator} to provide feedback for atomic agents to adjust their responses. For example, "The code should use a smaller batch size to reduce memory usage and employ TensorRT to accelerate on-device inference". As such, the grading process for response $r_i$ is expressed as:
\begin{equation}
    \{s_l^i, f_l^i, s_g^i, f_g^i\}=\mathcal{L}(r_i\ \vert\ \mathcal{I}_i,\mathcal{A}_i, u, \mathcal{S})
\end{equation}
where $s_l^i$ and $s_g^i$ are the local and global semantic scores, $f_l^i$ and $f_g^i$ are the corresponding feedback, respectively. $\mathcal{I}_i$ is the \textit{coordinator}'s instruction. $\mathcal{A}_i$ is the atomic agent's activity record during operation execution. $u$ is the user requirement. $\mathcal{S}$ is the runtime semantics obtained from the probe. The scores and dialogue will be stored in a dataset $D_a$:
\begin{equation}
\label{eq:atomic_agent_dataset}
    D_a=\{(\mathcal{A}_1, s_l^1, s_g^1), (\mathcal{A}_2, s_l^2, s_g^2), \cdots\},
\end{equation}
which will be further used for continuous tuning (\S~\ref{sec:continuous_LLM_tuning}).

\noindent\textbf{Response Adjustment}.
Considering both the current state and the feedback from the \textit{coordinator}, we prompt the atomic agent to adjust its response and regenerate a better one: $\hat{r_i} = \mathcal{L}(\mathcal{A}_i \ \vert \ (f_l^i, f_g^i))$. Note that we do not request agents to regenerate if the semantic score exceeds four, as the response is sufficiently aligned with the intended task, exhibiting only minor flaws that do not impact overall coherence \cite{wang2022self}.

\noindent\textbf{Remarks.}
We leverage the \textit{coordinator} to provide guidance grounded in mobile devices' runtime constraints, enabling atomic agents to adjust responses that better align with both the current operation and the overall user requirement. 

\subsection{Continuous LLM Tuning Paradigm}
\label{sec:continuous_LLM_tuning}

This module performs \textbf{self-evolution} by continuously tuning all the agents to enhance their reasoning about evolving edge AI knowledge.
We focus on three aspects of enhancement: 1) The \textit{coordinator}'s cognitive reasoning to autonomously select the next appropriate operation. 2) Atomic agents' operation execution abilities. 3) The \textit{programmer}'s domain-specific reasoning abilities to interpret edge AI knowledge and generate relevant algorithms.
Though building a RAG database \cite{gao2023retrieval} is promising, its 
timely maintenance demands substantial domain expertise, and vanilla LLMs still require strong reasoning abilities to comprehend the new knowledge \cite{shen2025gpiot}. In contrast, we adopt a \textit{modified} reinforcement fine-tuning (RFT) \cite{RFT} strategy to enhance agents' reasoning abilities in edge AI without user intervention, which 
can enhance LLMs' reasoning capabilities in specific domains by tuning on smaller amounts of data.

\noindent\textbf{RFT Procedure.}
Unlike supervised fine-tuning \cite{zhang2025dissecting, liu2025cronus, wan2025srpo} that forces LLMs to generate precise answers, RFT relies on a grader that scores generated responses, followed by policy gradients to refine LLM parameters \cite{yuan2022general}. RFT mainly consists of three stages: 
1) \textit{Grader design.}
A grader is created to score the quality of LLM outputs. Generally, graders can take various forms, such as format checkers, Python code, or even LLMs \cite{RFT}.
2) \textit{Dataset construction.} 
We first prepare a question set, $Q$, with an expected answer set, $A_e$, and input $Q$ into the LLM, $\mathcal{L}$, to obtain a generated answer set, $A_g$. Accordingly, we prompt the grader, $\mathcal{G}$, to score each answer. The tuning dataset, $T$, is thus obtained as a set of questions, expected answers, generated answers, and scores, $s_i$:
\begin{equation}
\begin{array}{l}
     a_i^g=\mathcal{L}(q_i), \quad \forall q_i\in Q\\
     s_i=\mathcal{G}(q_i,a_i^g, a_i^e),\quad \forall a_i^g\in A_g,\; \forall a_i^e\in A_e\\
     T=\{(q_i, a_i^g, a_i^e,s_i), \cdots\}
\end{array}
\end{equation}
3) \textit{Fine-tuning.}
With the obtained training dataset, policy-gradient updates (\eg, VPG \cite{yuan2022general}) are adopted for LLM tuning:
\begin{equation}
    \hat{g} = \frac{1}{\vert T\vert} \sum_{s_i \in T} \nabla_{\theta} \log s_i \big|_{\theta} \hat{A},\quad \theta^\prime=\theta+\eta\cdot\hat{g}
\end{equation}
where $\hat{g}$ is the policy gradient, $\vert T\vert$ is the dataset cardinality, $\theta$ are LLM parameters, $\hat{A}$ are the advantage estimates \cite{schulman2015high} based on the grader, and $\eta$ is the learning rate. After several epochs, RFT enhances the reasoning capabilities of LLMs by encouraging them to generate high-scoring outputs more frequently while suppressing low-scoring ones.

\noindent\textbf{Challenges.}
Directly adopting RFT in \project faces two main challenges: 
1) Assessing the continuously evolving \textit{coordinator} and atomic agents necessitates multiple evolving graders.
2) Grading the \textit{programmer} demands complicated analysis of both the quality and performance of the generated code across diverse edge AI algorithms. To tackle these issues, we propose: 1) An \textit{ensemble mutual-actor-critic strategy} to enhance both the \textit{coordinator} and atomic agents in a mutual-critique manner. Our insight is that both the \textit{coordinator} and atomic agents can function as graders to criticize each other's behaviors from distinct perspectives. 2) A \textit{self-reflective grader} to further critique the \textit{programmer}, considering both code performance and algorithm specialization to the edge AI domain.

\begin{figure}[t]
    \centering
    \includegraphics[width=0.48\textwidth]{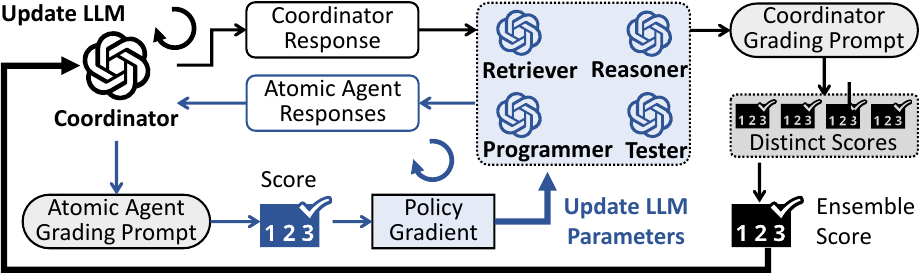}
    \caption{Ensemble mutual-actor-critic strategy.}
    \label{fig:ensemble_actor_critic}
\end{figure}

\begin{cmt}{\textit{Coordinator} Grading Prompt -- by \textit{retriever} (simplified)}{}
Given the \textbf{coordinator}'s response (the atomic operation and the instruction), please rate it based on the following two questions:\\
1. \textit{If the atomic operation is \hl{"retrieve"}, to what extent does it align with the current context?} For example, if the \textit{coordinator} proposes retrieving the same information (\eg, dataset specs) again, the score may be lower; otherwise, it may be higher if the information is important.\\
2. \textit{If not, how likely is it that this should be a \hl{"retrieve"} operation?}
\end{cmt}

\noindent\textbf{Ensemble Mutual-Actor-Critic Strategy.}
We augment the actor-critic algorithm \cite{konda1999actor} with an ensemble updating strategy to refine both the \textit{coordinator} and atomic agents. As shown in Fig.~\ref{fig:ensemble_actor_critic}, to update atomic agents, we treat the \textit{coordinator} as the grader to criticize atomic agent behaviors. In turn, to refine the \textit{coordinator}, we regard atomic agents as four separate graders, each criticizing the \textit{coordinator} from a distinct angle regarding their specific tasks. Thus, both the \textit{coordinator} and atomic agents can be mutually updated as self-exploration progresses.
Given the dataset $D_a$ in Equ.~\ref{eq:atomic_agent_dataset} containing the scores for the atomic agents, their tuning processes are expressed as:
\begin{equation}
    s_i^a=\alpha\cdot s_i^l + \beta\cdot s_i^g,\;\hat{g} = \frac{1}{\vert D_a\vert} \sum{\nabla_{\theta} \log s_i^a \big|_{\theta} \hat{A}},\; \theta^\prime=\theta+\eta\cdot\hat{g}
\end{equation}
where $s_i^a$ is a linear combination of both local and global semantic scores graded by the \textit{coordinator}. To tune the \textit{coordinator}, we first design four tailored grader prompts for atomic agents to rate the current atomic operation and instruction generated by the \textit{coordinator}. Taking the \textit{retriever} as an example, the grading prompt (\hl{P.4}) considers both the correctness and necessity of the \textit{coordinator}'s decision. An ensemble score is then calculated via weighted aggregation:
\begin{equation}
    \begin{array}{l}
         s_i^c=\sum_{o\in\mathcal{O}}{w_o \cdot \rho_o}, \quad s_i^c\in[0,5]\\
         \rho_o=\left\{\begin{array}{ll}
            s_i^o & \text{if the \textit{coordinator}'s operation = }o \\
            5 - s_i^o  & \text{otherwise}
         \end{array}\right.
    \end{array}
\end{equation}
where $o$ is the current operation, $w_o$ is its assigned weight, and $s_i^o$ is the score rated by each atomic agent. If $o$ does not match the atomic agent's operation, we compute the score as $5 - s_i^o$, as in this case, a higher score means that the atomic agent considers its own operation to be more appropriate. The \textit{coordinator} is then updated using the ensemble score:
\begin{equation}
    \quad\hat{g} = \frac{1}{\vert D_c\vert} \sum{\nabla_{\theta} \log s_i^o \big|_{\theta} \hat{A}},\quad \theta^\prime=\theta+\eta\cdot\hat{g}
\end{equation}
Such an ensemble strategy allows bidirectional critique between the \textit{coordinator} and atomic agents during self-exploration, thereby fostering reflection on accumulated context and enabling continuous, iterative self-evolution.


\noindent\textbf{Self-Reflective Grader for Edge AI.}
To further enhance the \textit{programmer}, we consider three main aspects:

\begin{itemize}[leftmargin=9pt]
    \item \textit{Code quality} includes syntax correctness and code smells, which are bad coding styles (\eg, no exception handlers) or potential weaknesses (\eg, deprecated packages) \cite{azeem2019machine}. They can be scored by SOTA code quality inspectors.
    \item \textit{Code performance} is tested by prompting the \textit{tester} to produce test cases and assess performance across multiple evaluation metrics. The \textit{tester} is further prompted to assign a score to comprehensively reflect the results.
    \item \textit{Algorithm specialization}: Whether the \textit{programmer} uses tailored algorithms to process sensor data. For instance, Butterworth filtering is commonly applied to Wi-Fi CSI data to suppress noise induced by non-human activities.
\end{itemize}

\noindent We prompt the \textit{programmer} to first identify a list of sensor data processing algorithms used in the generated code. For each algorithm, we prompt (\hl{P.5}) the \textit{programmer} itself to grade it on a scale from 1 (completely mismatched) to 6 (fully tailored) based on how closely the algorithm aligns with established, modality-specific methods. Higher scores indicate stronger adherence to domain-relevant practices, and lower scores reflect reliance on generic or unsuitable approaches. This process exploits the \textit{self-reflection} capability of the \textit{programmer}, with the reflection score further used in RFT to facilitate self-evolution. Then the final score is:
\begin{equation}
    s_i^p=s_i^q+s_i^{cp}+\sum{\lambda_j\cdot s_i^{s_j}},
\end{equation}
where $s_i^q$ is the code quality score, $s_i^{cp}$ is the code performance score, $s_i^{s_j}$ is the $j$-th algorithm specialization score with corresponding weight $\lambda_j$. Note that all tuning is performed entirely offline, without interfering with the online code generation.

\begin{figure*}[t]
    \centering
    \subfigure[Code embedding similarity]{
        \includegraphics[width=0.225\textwidth]{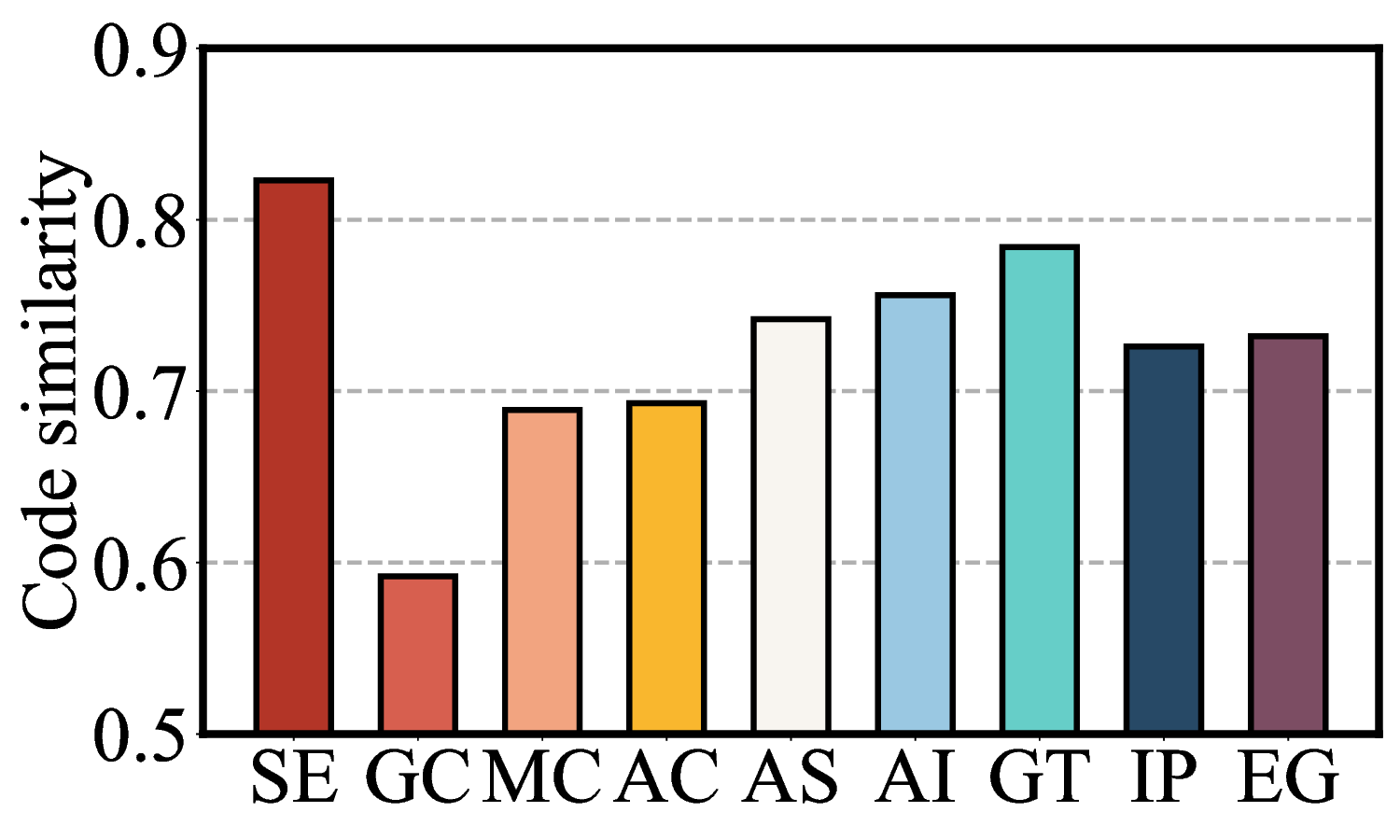}
        \label{fig:benchmark_code_similarity}
    }
    \centering
    \subfigure[Pass@1]{
        \includegraphics[width=0.225\textwidth]{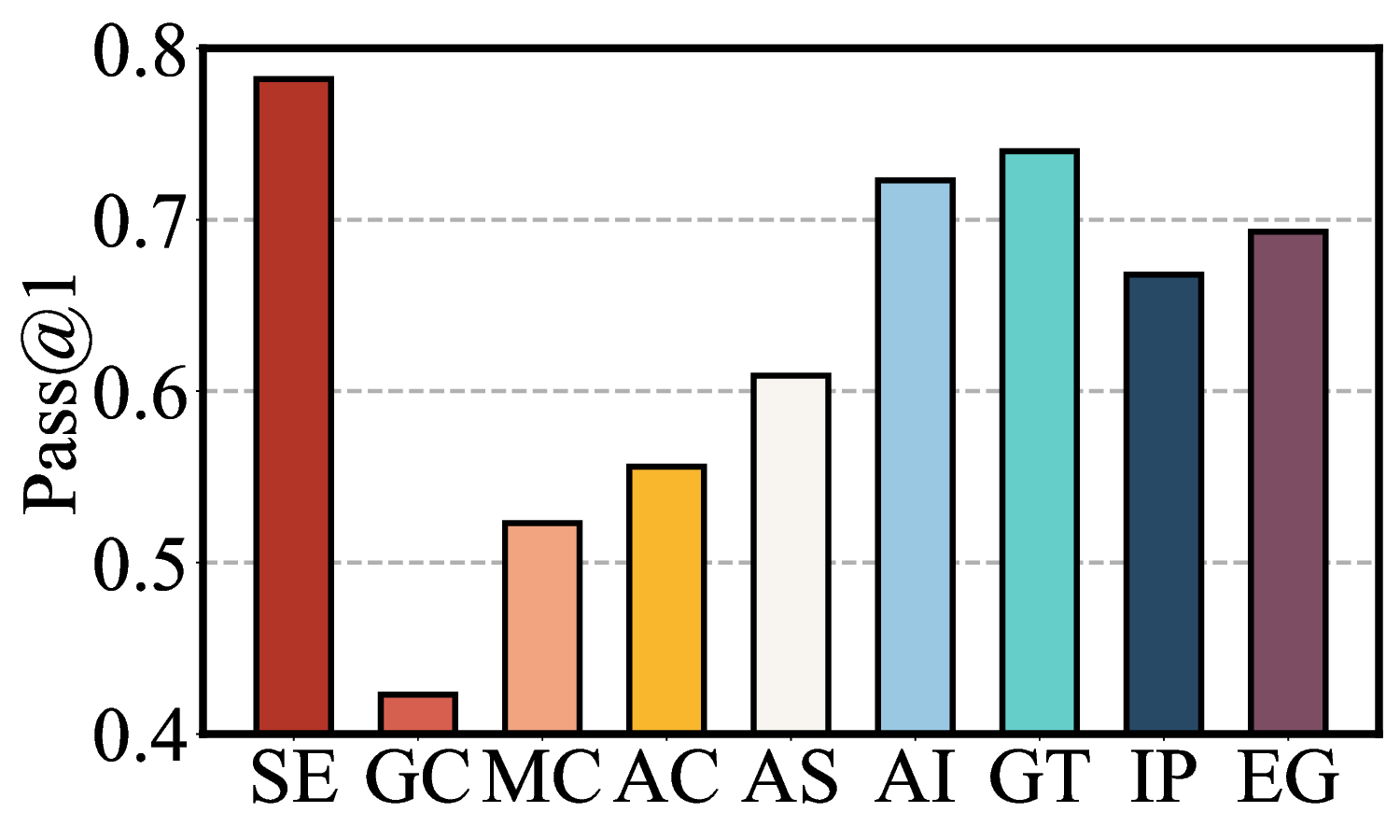}
        \label{fig:benchmark_pass1}
    }
    \centering
    \subfigure[Pass@5]{
        \includegraphics[width=0.225\textwidth]{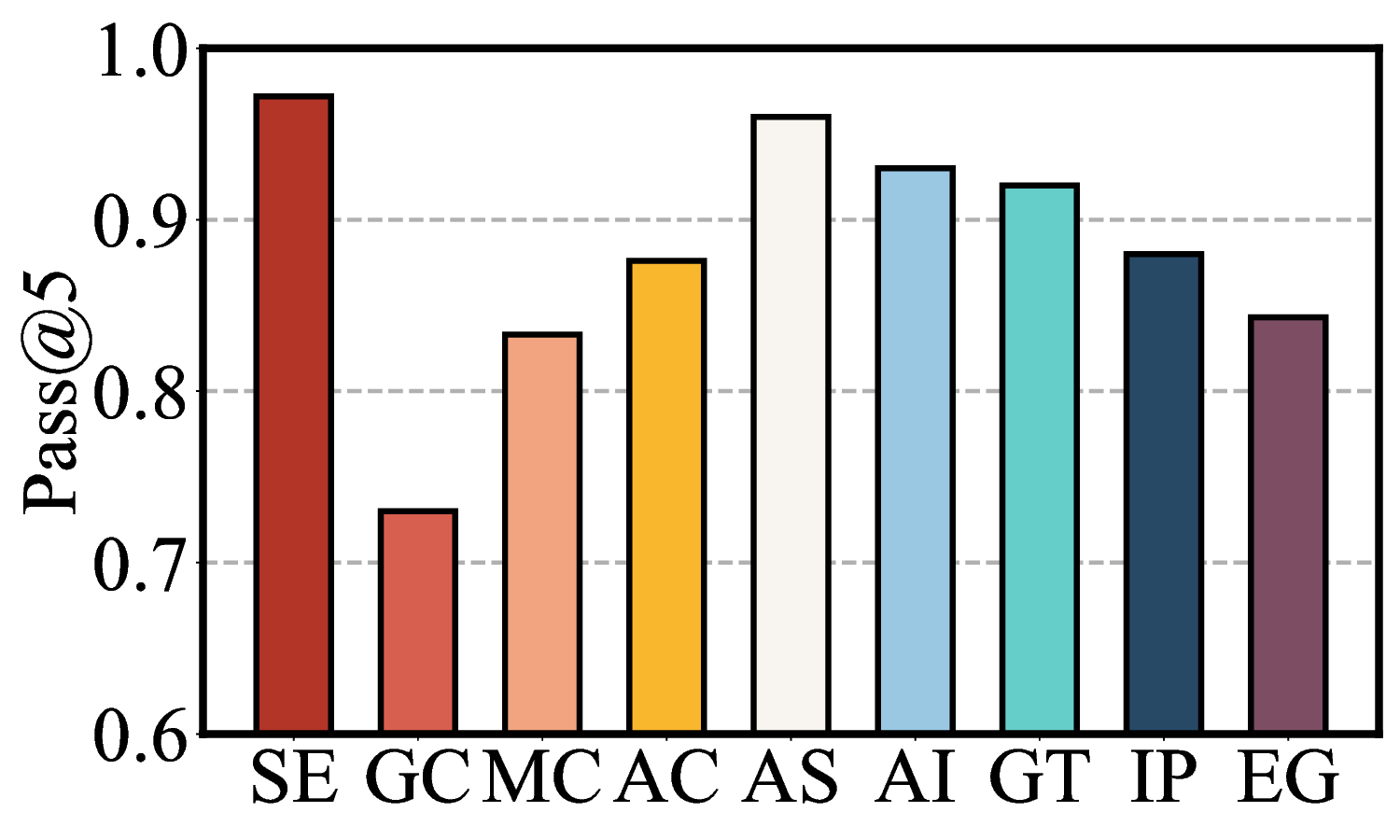}
        \label{fig:benchmark_pass5}
    }
    \centering
    \subfigure[Code quality score]{
        \includegraphics[width=0.225\textwidth]{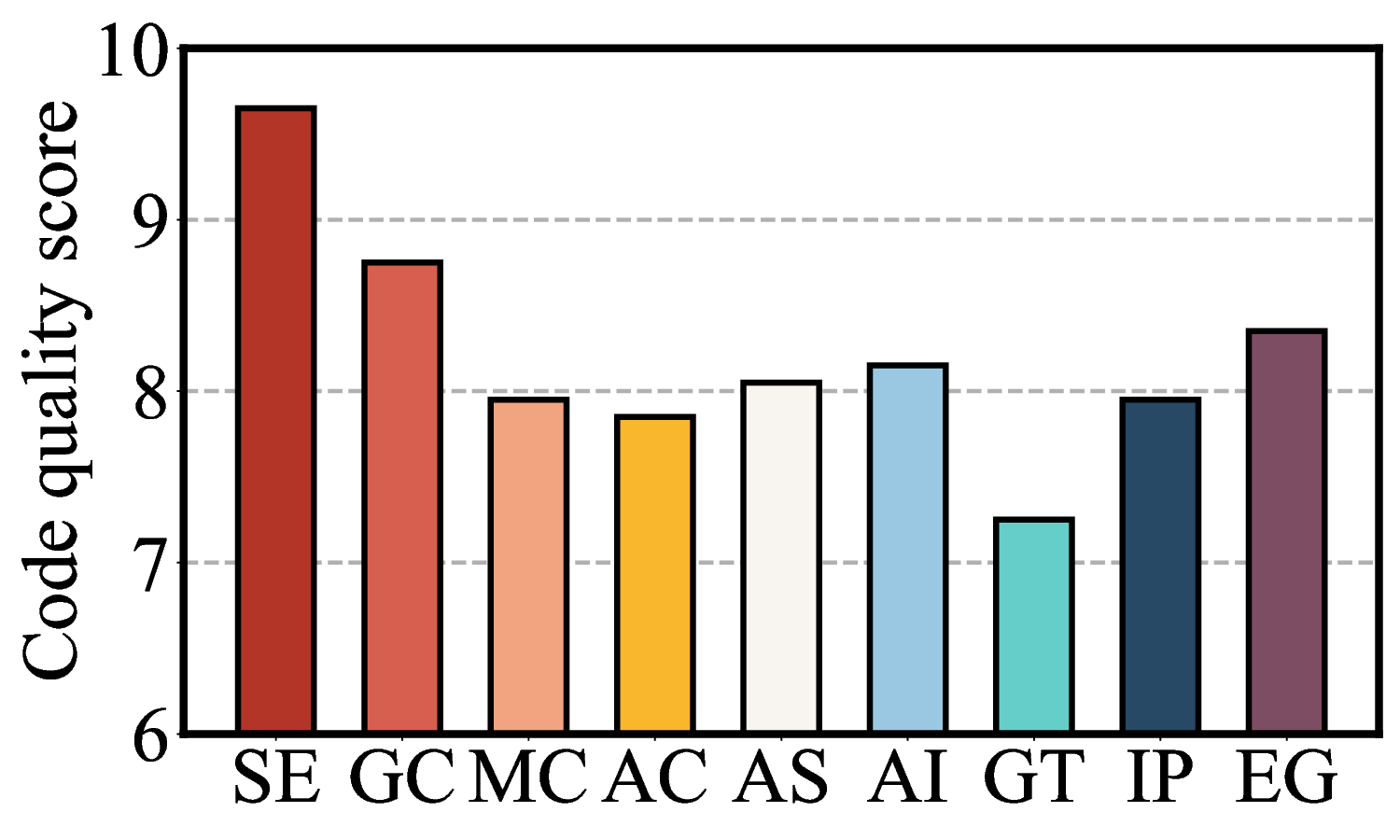}
        \label{fig:benchmark_code_quality}
    }
    \caption{Performance comparison on \textit{IoTBench} (SE: \project, GC: GitHub Copilot, MC: MapCoder, AC: AgentCoder, AS: AgentSquare, AI: AutoIOT, GT: GPIoT, IP: IoTPilot, EG: AutoEmbed).}
    \label{fig:performance_comparison_benchmark}
\end{figure*}

\begin{cmt}{Algorithm Specialization Grading Prompt (simplified)}{}
You are a \textbf{code auditor} tasked with scoring how well the algorithm is tailored to the target data modality, rather than using generic methods.\\
\vspace{-8pt}
\raisebox{1.8ex}{\rule{0.7\textwidth}{0.5pt}}\\
\textbf{Tailoring score} (1–6): \\
6 -- \textit{Fully}: Domain algorithms with modality-appropriate preprocessing.\\
5 -- \textit{Mostly}: Domain algorithm present with minor omissions.\\
4 -- \textit{Partially}: Some domain steps present, but key elements are missing.\\
3 -- \textit{Weakly}: Mostly generic with possible accuracy degradation.\\
2 -- \textit{Not tailored}: Generic algorithm; lacks essential modality logic.\\
1 -- \textit{Mismatched/harmful}: Algorithm contradicts modality requirements.
\end{cmt}

\section{Experiment Setup}
\subsection{Implementation}
We fully implement \project using o4-mini \cite{o4_model} as the internal LLM for balanced reasoning and response latency. The initial tool pool contains a Tavily-based search engine \cite{Asher2022tavily}, a Chroma-based database \cite{Bashir2022chroma}, a Python executor inside a sandbox, a Pylint-based code quality inspector \cite{Pylint_contributors_Pylint}, and a tool builder that allows agents to dynamically create new tools. \textit{We carefully design all graders and adopt RFT following OpenAI's official guidelines} \cite{RFT}. We develop \project with 5,288 lines of Python code and deploy it on a local server. The generated code is tested and deployed on various mobile devices, including portable laptops, Jetson Nano, and Raspberry Pis, which are connected to the local server via a router.

\subsection{Evaluation Protocols}
We use both a SOTA benchmark and three representative edge AI applications for comprehensive evaluations. The benchmark provides a controlled setting for measuring core capabilities, while the real-world edge AI cases investigate the system's effectiveness and adaptability in practical scenarios. In addition, we conduct a user study to evaluate the \textit{\textbf{generalizability}} of \project, where we invite users to propose diverse improvisational requirements (including unseen applications). 

\subsubsection{\textbf{Benchmark}}
We use \textit{IoTBench} \cite{shen2025gpiot}, a tailored benchmark to assess LLMs' abilities in generating IoT-relevant code. 
\textit{We filter out data irrelevant to edge AI and augment the rest to 200 coding tasks}.
Each task contains a "\textit{task specification} $\rightarrow$ \textit{edge AI code}" pair with curated test cases, each corresponding to a specific functional requirement in the specification. We randomly split the augmented benchmark into two parts: 20\% for tuning and 80\% for testing (RFT requires less tuning data). \textit{The tuning and testing data have completely different tasks.}

\begin{figure*}[t]
    \centering
    \subfigure[Execution success rate]{
        \includegraphics[width=0.225\textwidth]{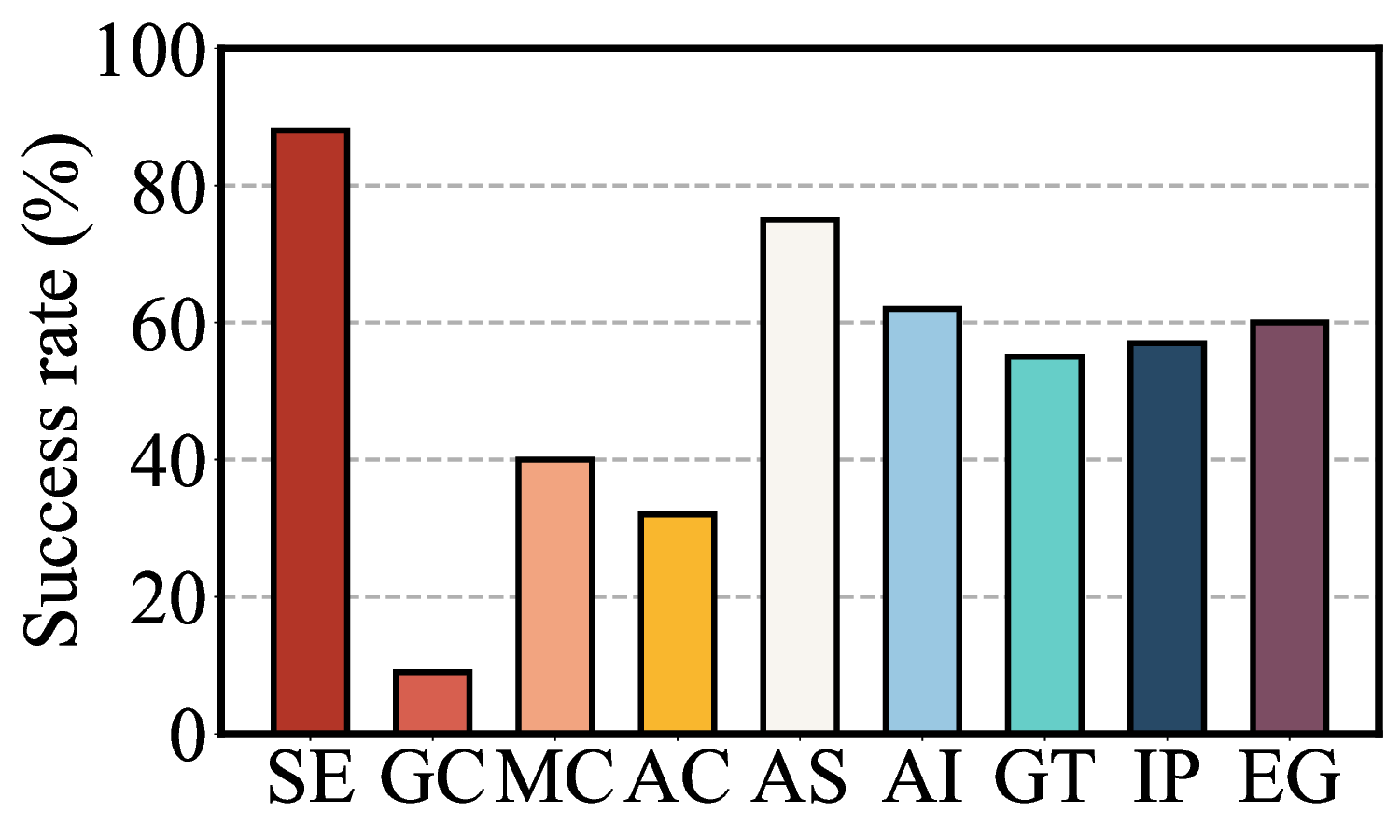}
        \label{fig:HAR_success_rate}
    }
    \centering
    \subfigure[Classification accuracy]{
        \includegraphics[width=0.225\textwidth]{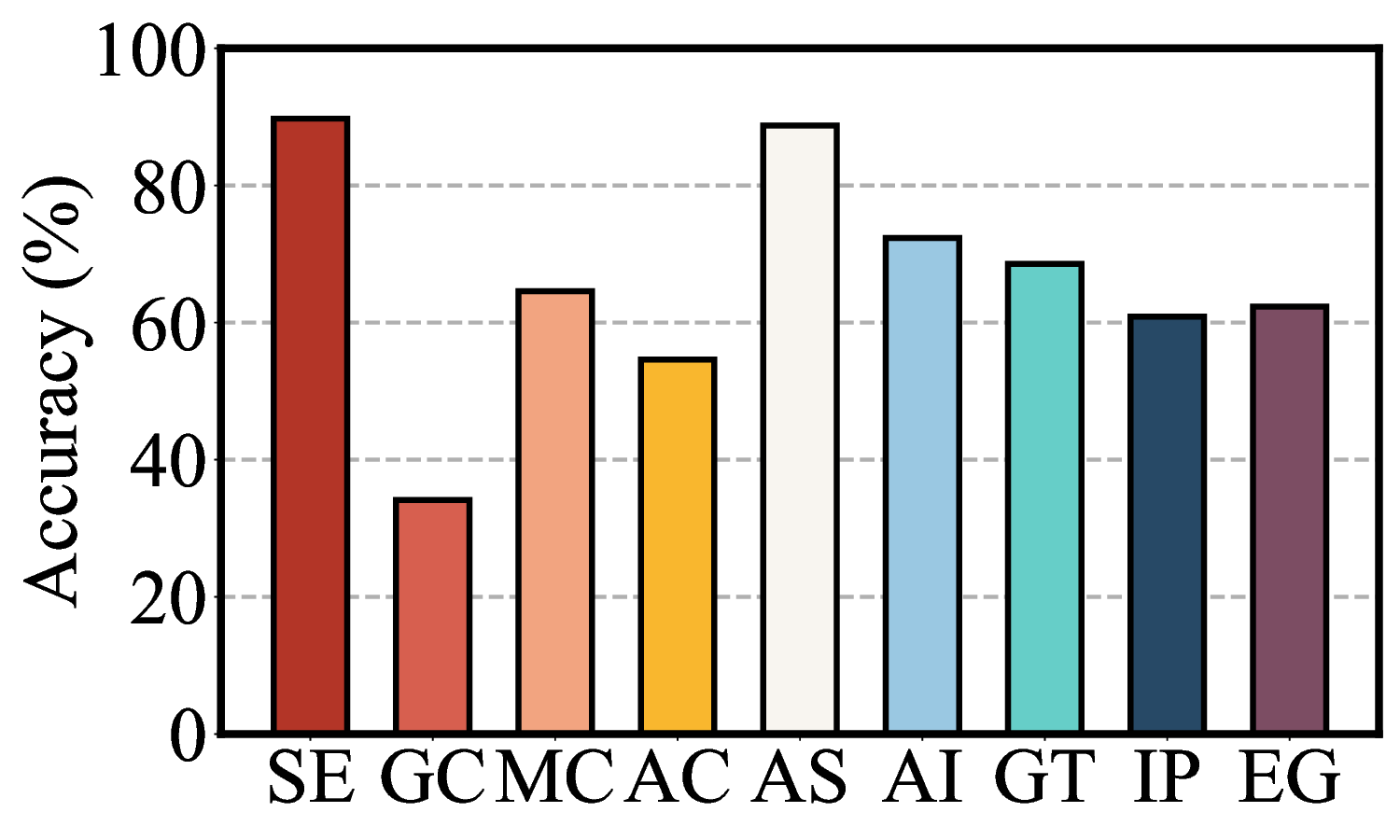}
        \label{fig:HAR_accuracy}
    }
    \centering
    \subfigure[Inference time \& memory usage]{
        \includegraphics[width=0.225\textwidth]{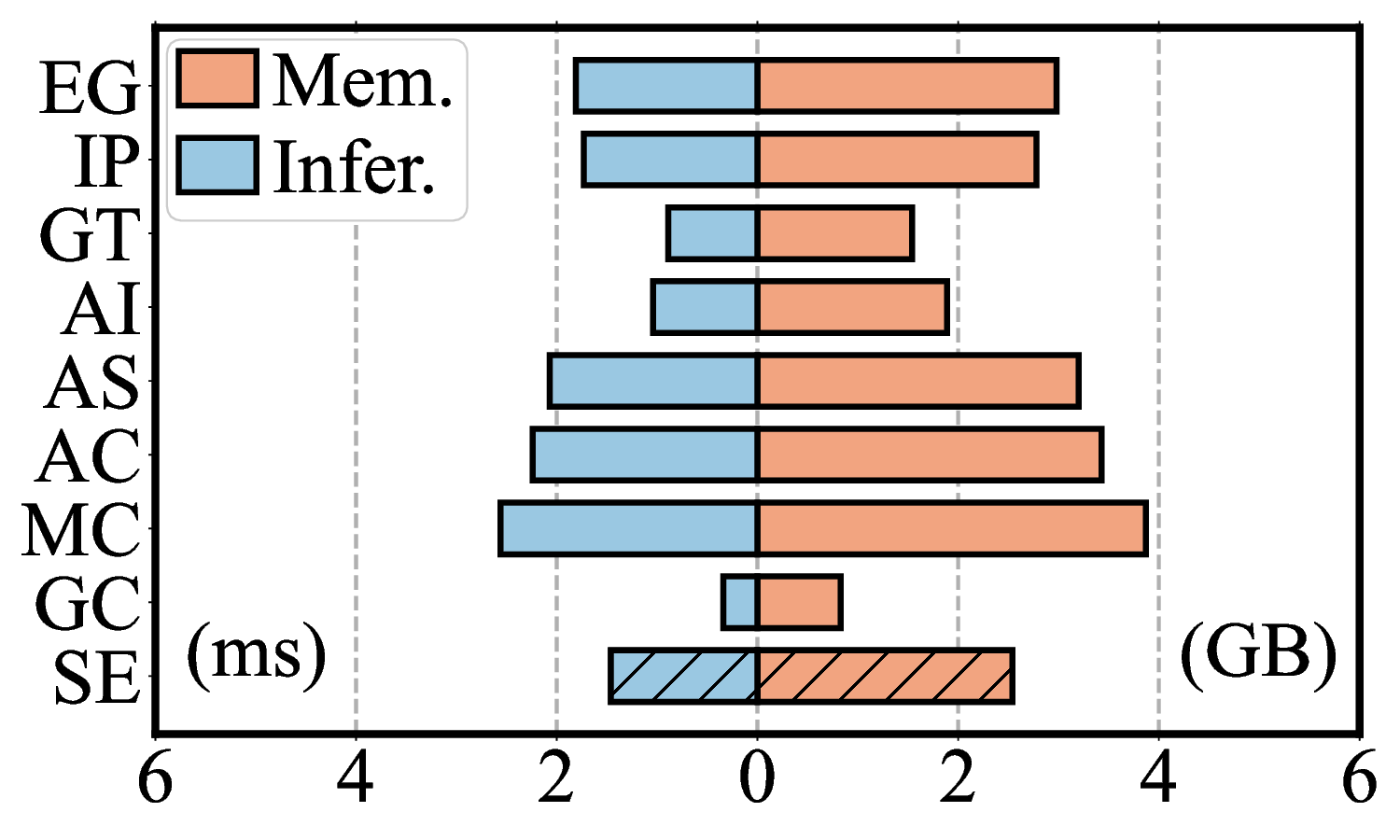}
        \label{fig:HAR_memory_inference}
    }
    \centering
    \subfigure[Wall-clock generation time]{
        \includegraphics[width=0.225\textwidth]{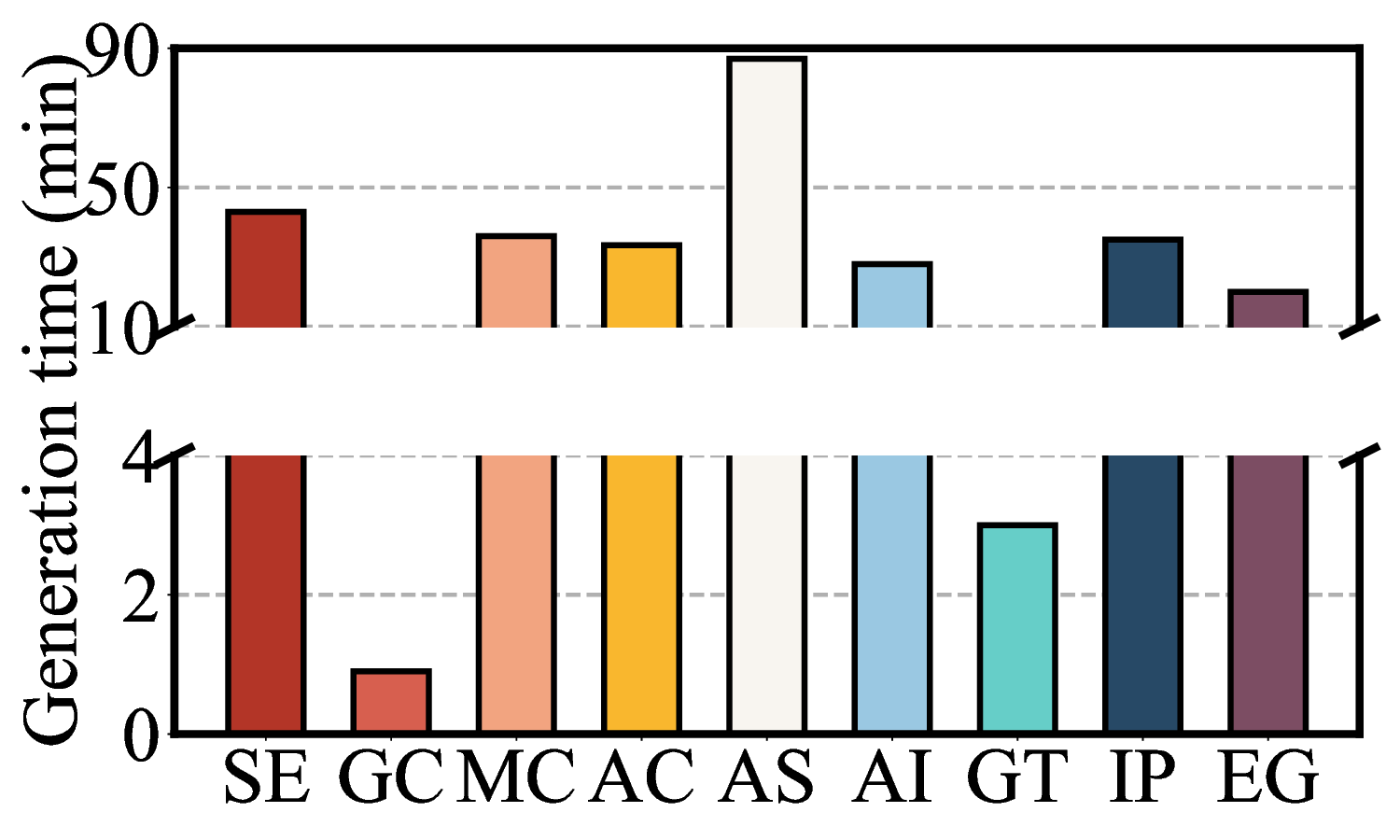}
        \label{fig:HAR_generation_time}
    }
    \caption{Performance comparison on HAR.}
    \label{fig:performance_comparison_HAR}
\end{figure*}

\subsubsection{\textbf{Multimodal Edge AI Applications}}
Considering different \textit{developing pipelines, data processing algorithms, sensor modalities, and device heterogeneity}, we select three representative multimodal edge AI applications for evaluation: 

\noindent\ding{182} \textbf{Human Activity Recognition (HAR)} using the XRF55 dataset \cite{wang2024xrf55}, containing 42.9K samples and 55 action classes of various human daily activities. Three modalities are involved: RFID, Wi-Fi CSI, and mmWave. \textit{We request deployment on a Raspberry Pi with no GPU and only 8GB RAM.} 

\noindent\ding{183} \textbf{Vision-Language Understanding (VLU)} using the MMMU dataset \cite{yue2024mmmu} for image Q\&A. There are two modalities in the dataset: image and text. \textit{We request deployment on a Jetson Nano board, which has a different runtime environment from the local server where \project operates.}

\noindent\ding{184} \textbf{Object Detection in Autonomous Driving (ODAD)} using the aiMotive dataset \cite{matuszka2022aimotive} in 176 scenes. This task is the most complex, as it involves four modalities (1 LiDAR, 2 radars, 4 cameras, and 1 GNSS+INS) with high dimensionality, necessitating advanced data processing algorithms and AI models. \textit{Besides, we randomly remove some data modalities to simulate the modality missing issue in real-world scenarios.}

\subsubsection{\textbf{Baselines.}}
We compare \project with 8 SOTA baselines:

\begin{itemize}[leftmargin=9pt]
    \item \textbf{General coding LLM}: \textit{GitHub Copilot} \cite{github} suggests code snippets and functions following user instructions.
    \item \textbf{General coding agents} (fixed workflow): \textit{MapCoder} \cite{islam2024mapcoder}, \textit{AgentCoder} \cite{huang2023agentcoder}, \textit{AgentSquare} \cite{shang2025agentsquare}.
    \item \textbf{Coding agents tailored to mobile tasks} (fixed workflow): \textit{AutoIOT} \cite{shen2025autoiot}, \textit{GPIoT} \cite{shen2025gpiot}, \textit{IoTPilot} \cite{gong2025programming}, \textit{AutoEmbed} \cite{yang2026autoembed}.
\end{itemize}

\section{Evaluation}
\label{sec:evaluation}

\subsection{Benchmark Evaluation}
\subsubsection{\textbf{Metrics}}
\textit{Code similarity}: we convert code into embeddings that capture its internal logic via CodeT5+ \cite{wang2023codet5}. We compute the cosine similarity between the embeddings of the generated and the reference code (from IoTBench). A higher value indicates a stronger ability to generate edge-AI-related code. \textit{Pass@k}: measures the probability that at least one of the top-k generated programs passes all functional test cases \cite{chen2021evaluating}. A high value indicates \project accurately interprets tasks and passes more functional test cases from IoTBench. \textit{Code quality}: we use Pylint \cite{Pylint_contributors_Pylint} to assess code quality.

\subsubsection{\textbf{Performance Comparison}}
As shown in Fig.~\ref{fig:performance_comparison_benchmark}, we find:

\textbf{\project can effectively generate functionally correct and performant code tailored to edge AI following task specifications.} On average, \project achieves an 11\% higher code similarity score and a 17\% higher pass@1 rate than baselines. The gain in pass@5 rate (10\% $\uparrow$) is relatively small because it has higher fault tolerance than pass@1. 

\textbf{Interestingly, general coding agents (GitHub Copilot, MapCoder, AgentCoder) exhibit lower performance} in terms of code similarity (14\% $\downarrow$) and pass@1 rate (25\% $\downarrow$), indicating that general coding agents are incapable of generating high-quality edge AI programs. For instance, they all directly apply AI models on the raw Wi-Fi CSI data without dedicated processing (\eg, Butterworth filtering \cite{butterworth1930theory}) to remove noise induced by non-human activities in HAR.

\textbf{In contrast, IoT-tailored agents (AutoIOT, GPIoT, IoTPilot, AutoEmbed) perform relatively better}, owing to tailored designs such as RAG. In addition, AgentSquare achieves comparable pass@5 performance to \project, due to its iterative workflow optimization method. However, we find that it requires much more time to generate performant code (\S~\ref{sec:case_study}).

\textbf{\project significantly outperforms the baselines in code quality by an average of 1.6 points}. The reason is that our \textit{self-reflective grader} uses code quality as a grading metric. Thus, after RFT, the \textit{programmer} gains enhanced abilities to produce high-quality code while adopting edge-AI-relevant algorithms. Notably, GitHub Copilot achieves higher code quality scores than other baselines, as it uses tailored LLMs and prompts for high-quality code generation, whereas other agents typically rely on vanilla LLMs. \textit{In general, the superior performance of \project stems from its self-exploration abilities in adopting SOTA algorithms dedicated to various edge AI tasks.}

\subsection{Case Study on Edge AI Applications}
\label{sec:case_study}
\subsubsection{\textbf{Metrics}}
We repeat the test 100 times and calculate the average values of the following metrics: \textit{Classification accuracy} (HAR): the portion of the correctly classified activities. \textit{Answer correct rate} (VLU): the portion of the correctly answered image Q\&A pairs. \textit{Precision} (ODAD): the portion of the detected bounding boxes that are actually correct. \textit{Memory consumption}: the CPU RAM or GPU VRAM usage during code execution. \textit{Inference time}: the time used for processing a single multimodal data batch on the target device. \textit{Execution success rate}: the portion of the programs that can be successfully executed. \textit{Wall-clock generation time}: the time from the user's request to the final program synthesis.

\begin{figure*}[t]
    \centering
    \subfigure[Execution success rate]{
        \includegraphics[width=0.225\textwidth]{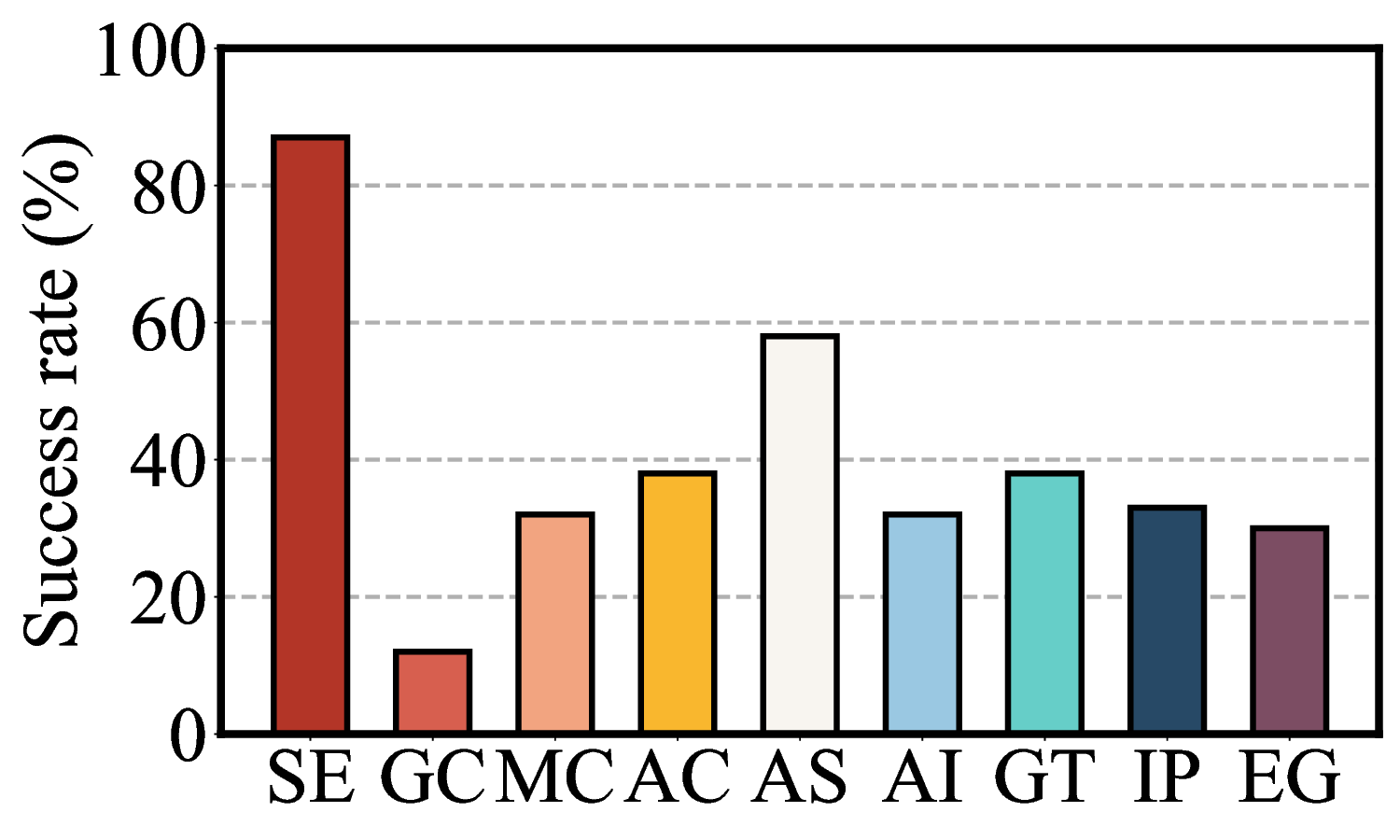}
        \label{fig:AD_success_rate}
    }
    \centering
    \subfigure[Answer correct rate]{
        \includegraphics[width=0.225\textwidth]{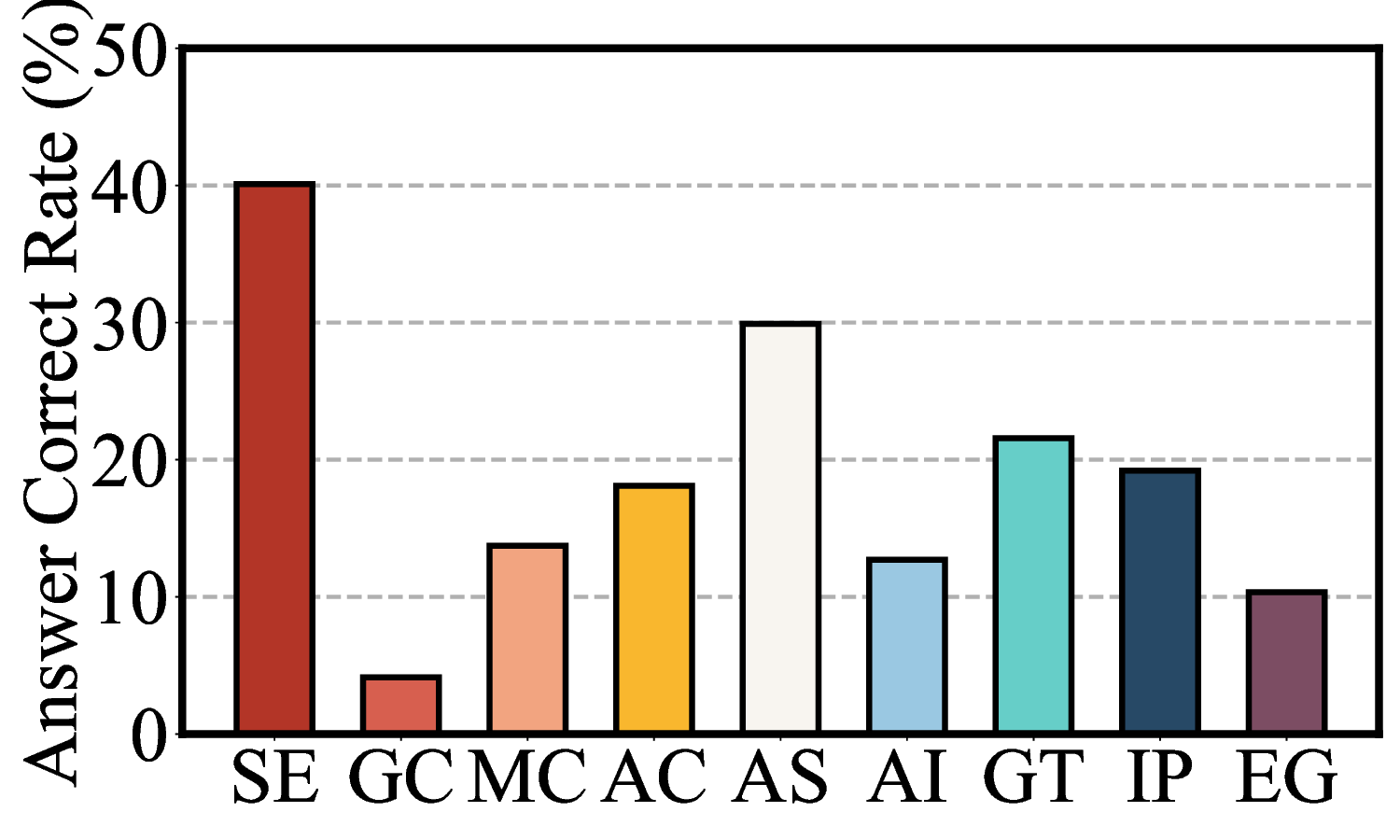}
        \label{fig:AD_accuracy}
    }
    \centering
    \subfigure[Inference time \& memory usage]{
        \includegraphics[width=0.225\textwidth]{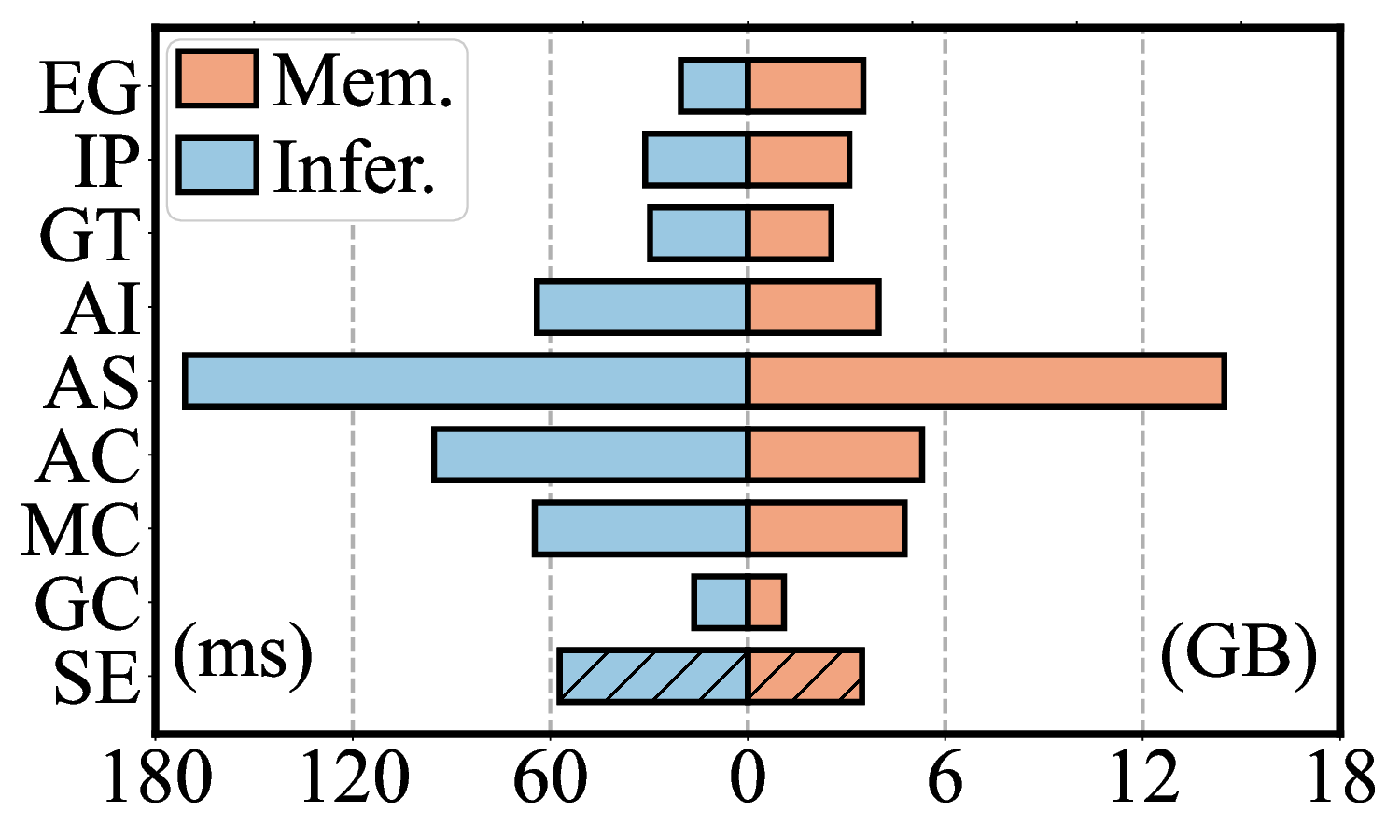}
        \label{fig:AD_memory_inference}
    }
    \centering
    \subfigure[Wall-clock generation time]{
        \includegraphics[width=0.225\textwidth]{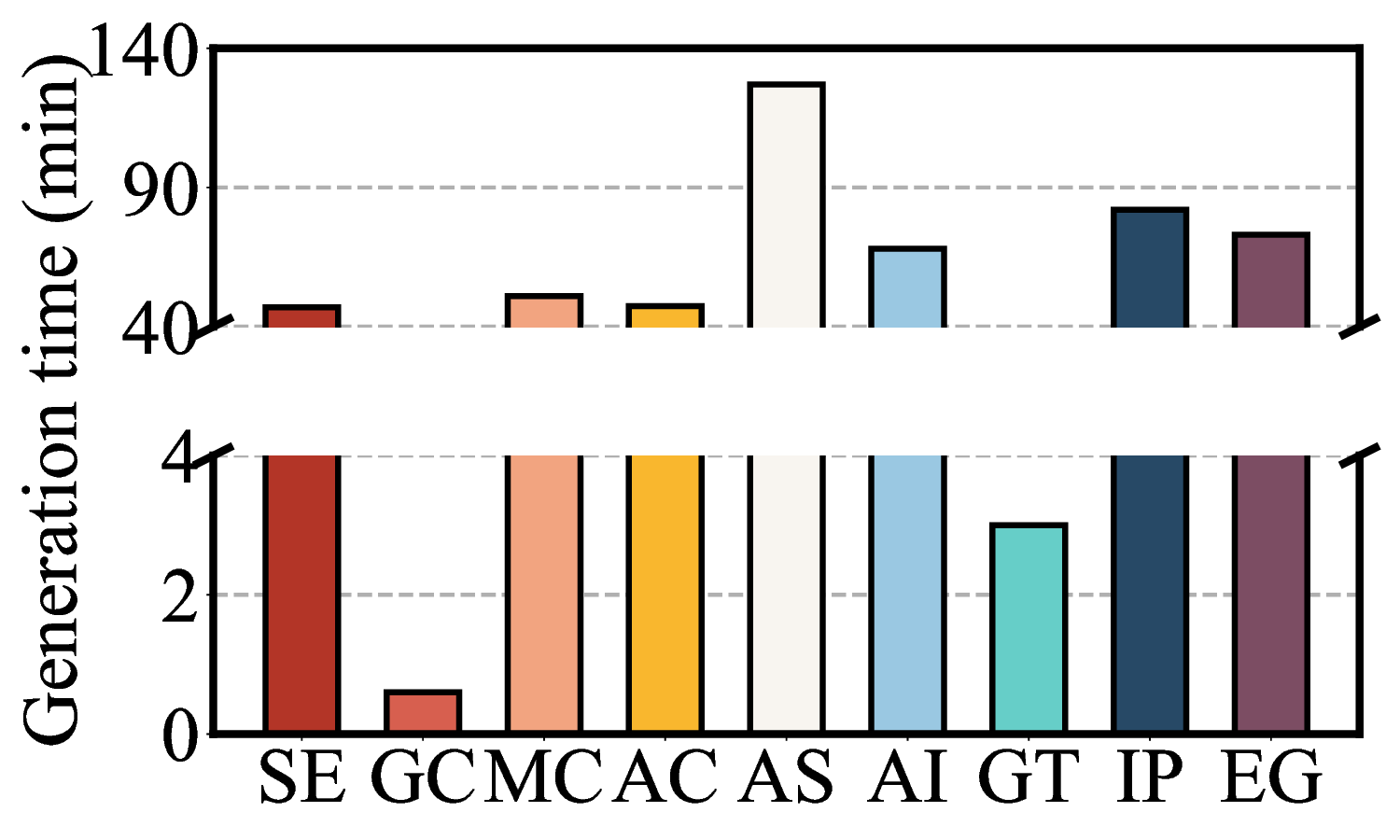}
        \label{fig:AD_generation_time}
    }
    \caption{Performance comparison on VLU.}
    \label{fig:performance_comparison_AD}
\end{figure*}

\begin{figure*}[t]
    \centering
    \subfigure[Execution success rate]{
        \includegraphics[width=0.225\textwidth]{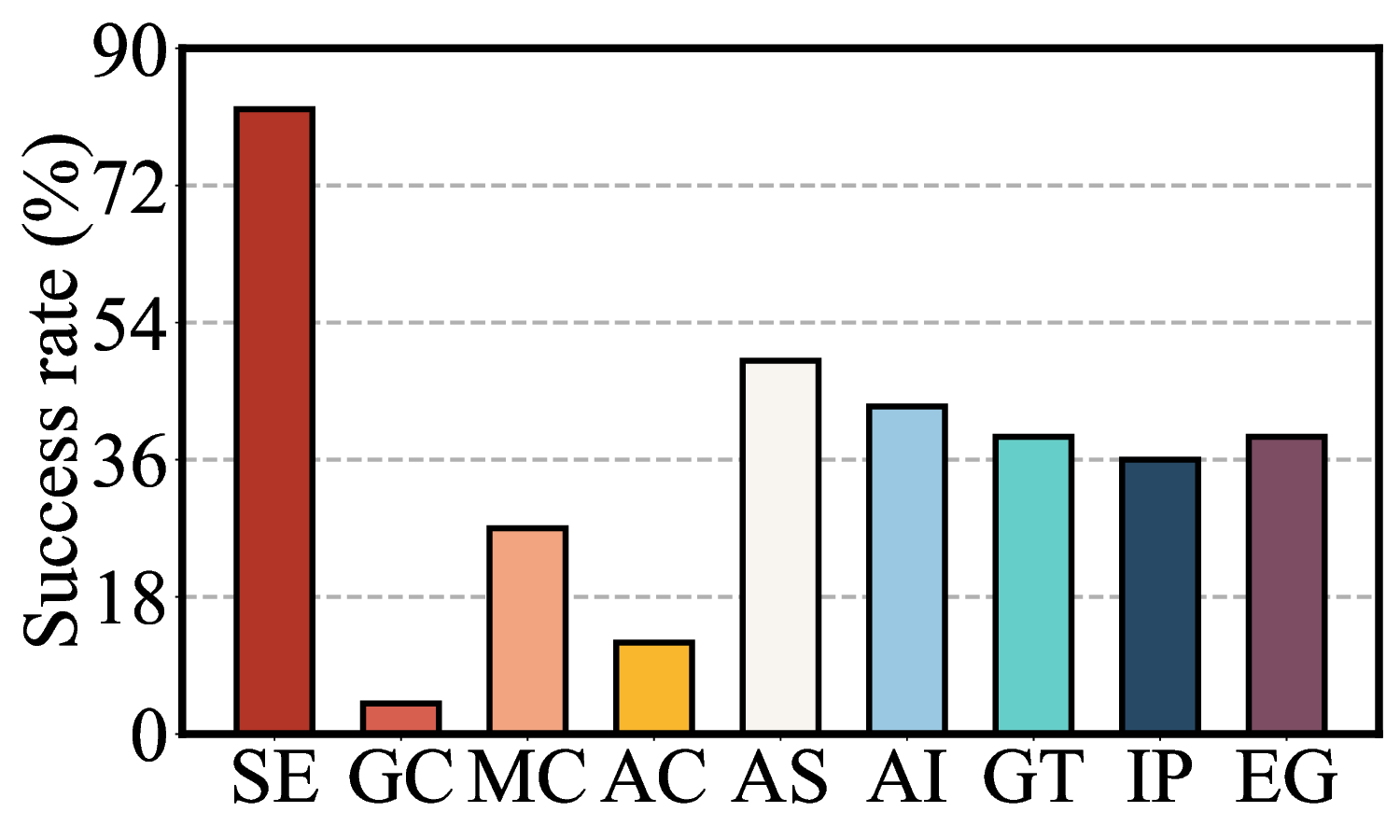}
        \label{fig:ODAD_success_rate}
    }
    \centering
    \subfigure[Average precision]{
        \includegraphics[width=0.225\textwidth]{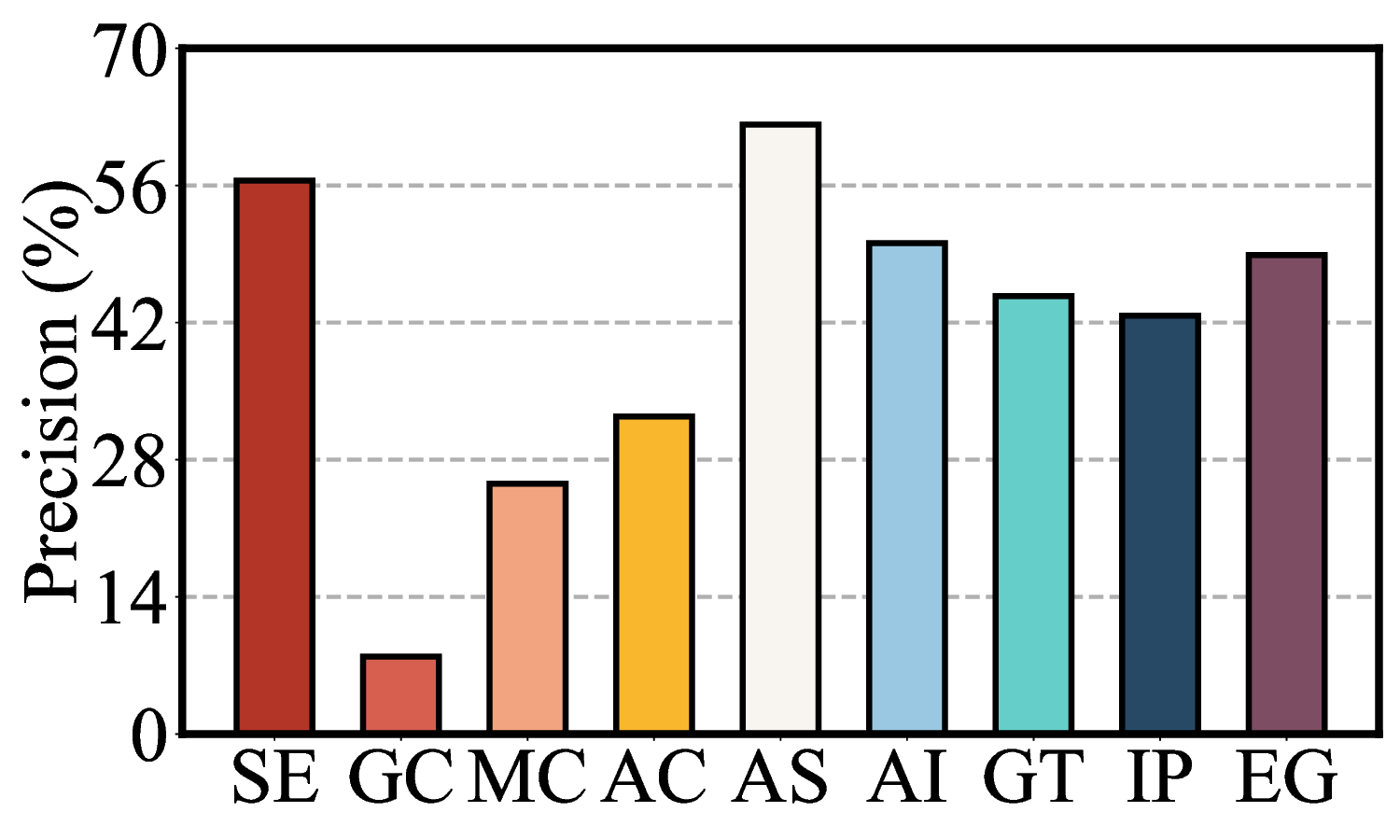}
        \label{fig:ODAD_precision}
    }
    \centering
    \subfigure[Inference time \& memory usage]{
        \includegraphics[width=0.225\textwidth]{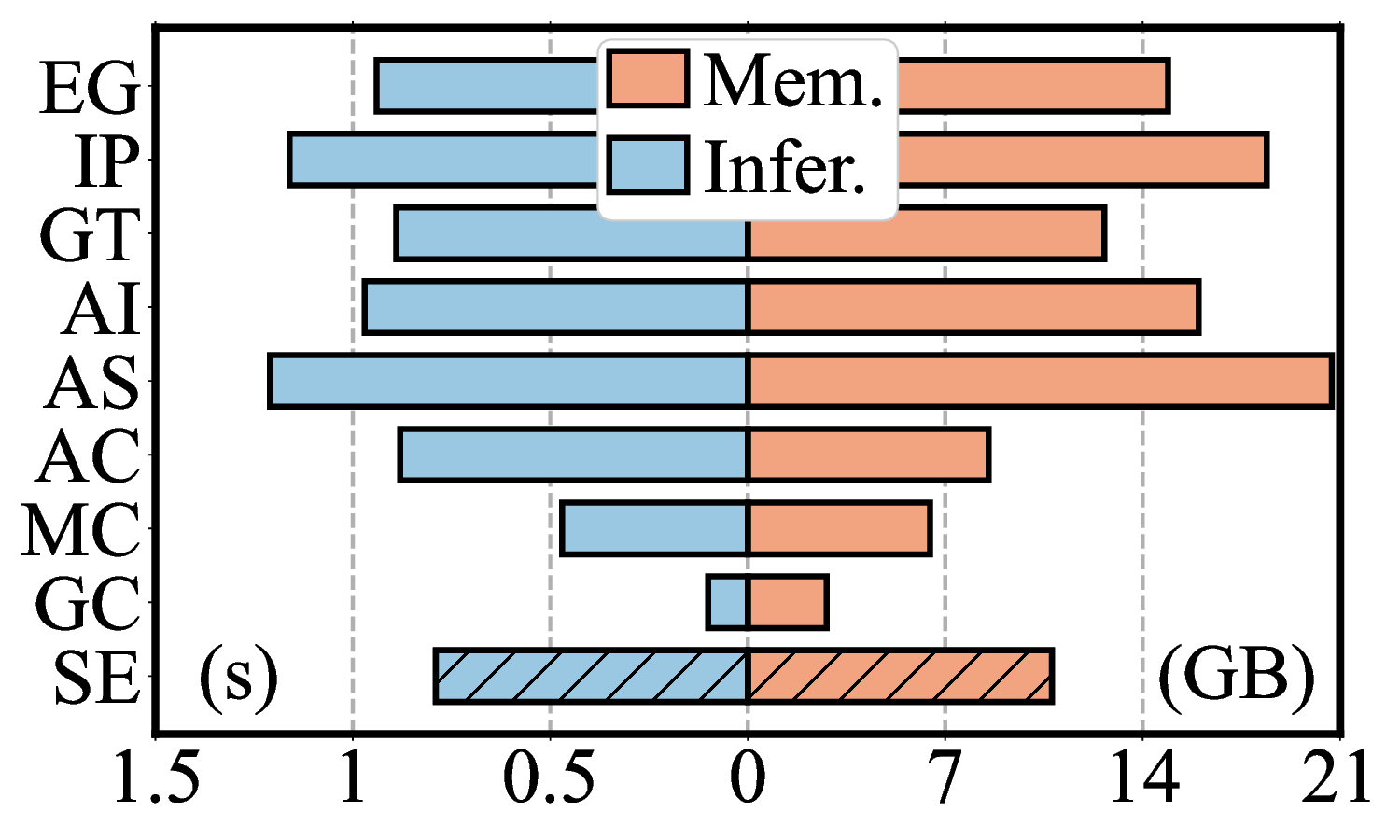}
        \label{fig:ODAD_memory_inference}
    }
    \centering
    \subfigure[Wall-clock generation time]{
        \includegraphics[width=0.225\textwidth]{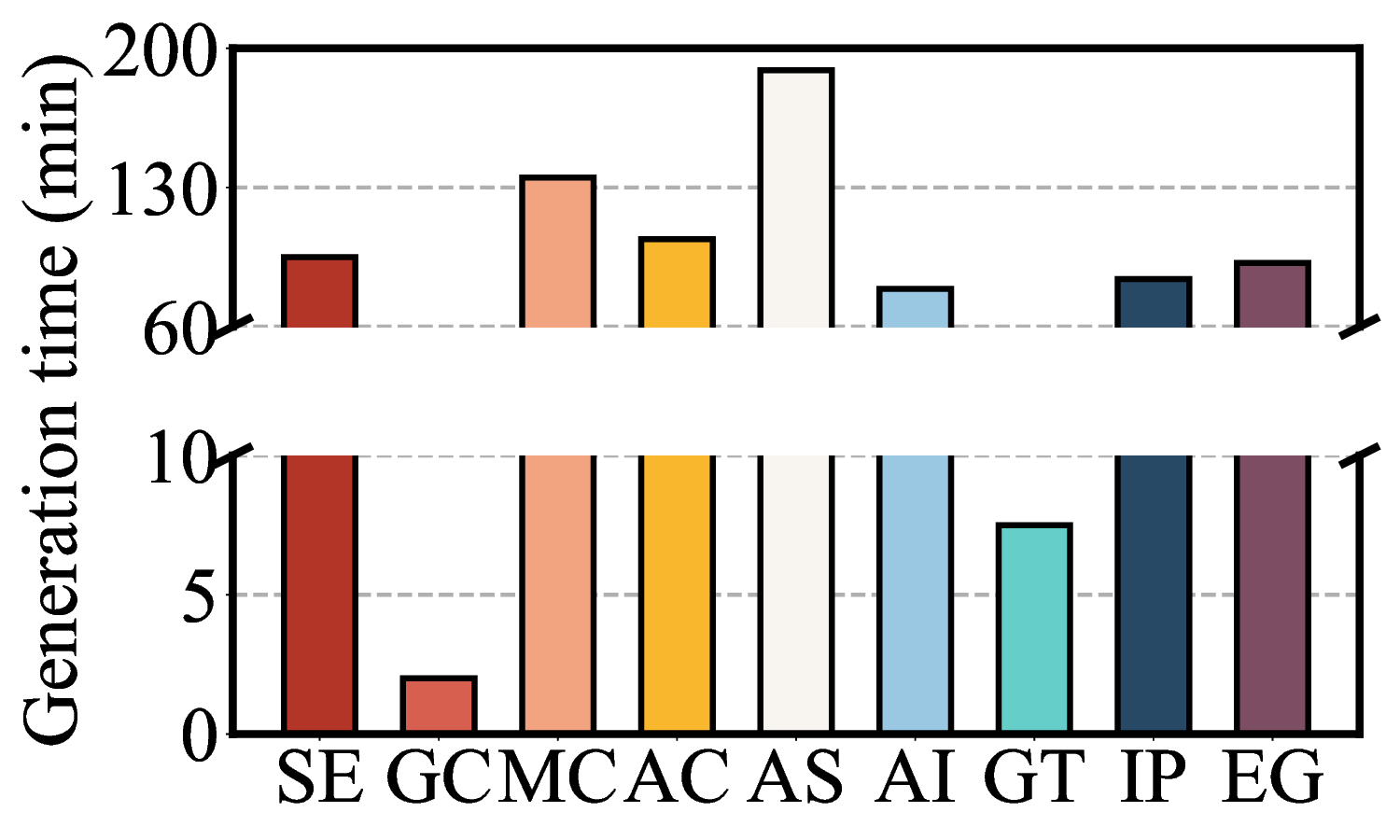}
        \label{fig:ODAD_generation_time}
    }
    \caption{Performance comparison on ODAD.}
    \label{fig:performance_comparison_ODAD}
\end{figure*}

\subsubsection{\textbf{Evaluation on HAR}}
\label{sec:HAR_evaluation}
We request that the multimodal HAR model be executed on a Raspberry Pi (8GB RAM, no GPU). As shown in Fig.~\ref{fig:performance_comparison_HAR}, \project outperforms baselines in terms of success rate (39\% $\uparrow$) and classification accuracy (26.5\% $\uparrow$). We find that \project first generates a script to obtain the data dimension, which is not specified in the requirement, indicating that \textit{\project learns to fetch such important yet missing information via self-exploration, rather than relying on explicit workflow engineering by developers.} In contrast, the baselines' fixed workflows often get stuck in handling unspecified data dimensions. Further, by decomposing the task into multiple subtasks, \project can independently generate, test, and refine code snippets for each, ultimately achieving higher accuracy.

Most GitHub-Copilot-generated code cannot be executed, while executable ones achieve only 34\% accuracy on average. The reason is that GitHub Copilot is designed for general-purpose coding and lacks specialized designs. Thus, it cannot effectively produce tailored edge AI code. This phenomenon is consistently observed across distinct edge AI tasks and is exacerbated as the task becomes complex (Fig.~\ref{fig:performance_comparison_AD} \& Fig.~\ref{fig:performance_comparison_ODAD}).

Moreover, \project-designed models consume moderate memory with acceptable inference latency (Fig.~\ref{fig:HAR_memory_inference}). 
Though AutoIOT and GPIoT adopt thorough model optimization methods (\eg, pruning \cite{DBLP:conf/mobicom/0005SLPLC21}) to reduce memory cost to 2GB, the resulting accuracy is much lower than \project. In addition, IoTPilot and AutoEmbed sometimes achieve lower performance, likely because they are more tailored to embedded tasks.

Lastly, \project only needs 43 minutes for code generation, slightly surpassing other agents that use fixed workflows. While GitHub Copilot and GPIoT take only a few minutes with a few rounds of LLM calls, most generated code is unexecutable. AgentSquare has a 13\% lower success rate and takes a much longer time (90 minutes) to generate executable code, as it exhaustively searches over large module spaces. 

\begin{figure}[t]
    \centering
    \subfigure[Runtime feedback mechanism]{
        \includegraphics[width=0.225\textwidth]{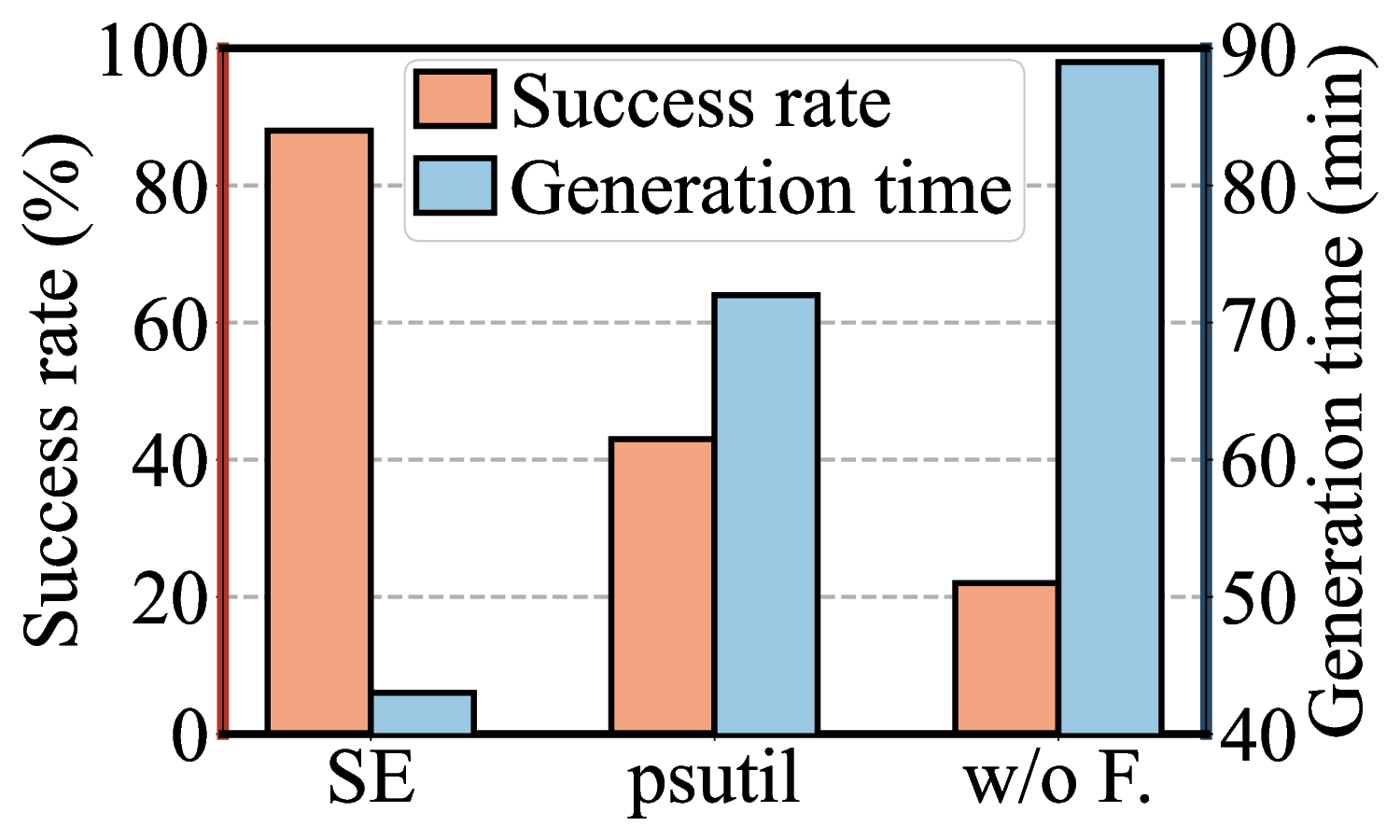}
        \label{fig:ablation_agentic_recovery_mechanism}
    }
    \centering
    \subfigure[Continuous LLM tuning]{
        \includegraphics[width=0.225\textwidth]{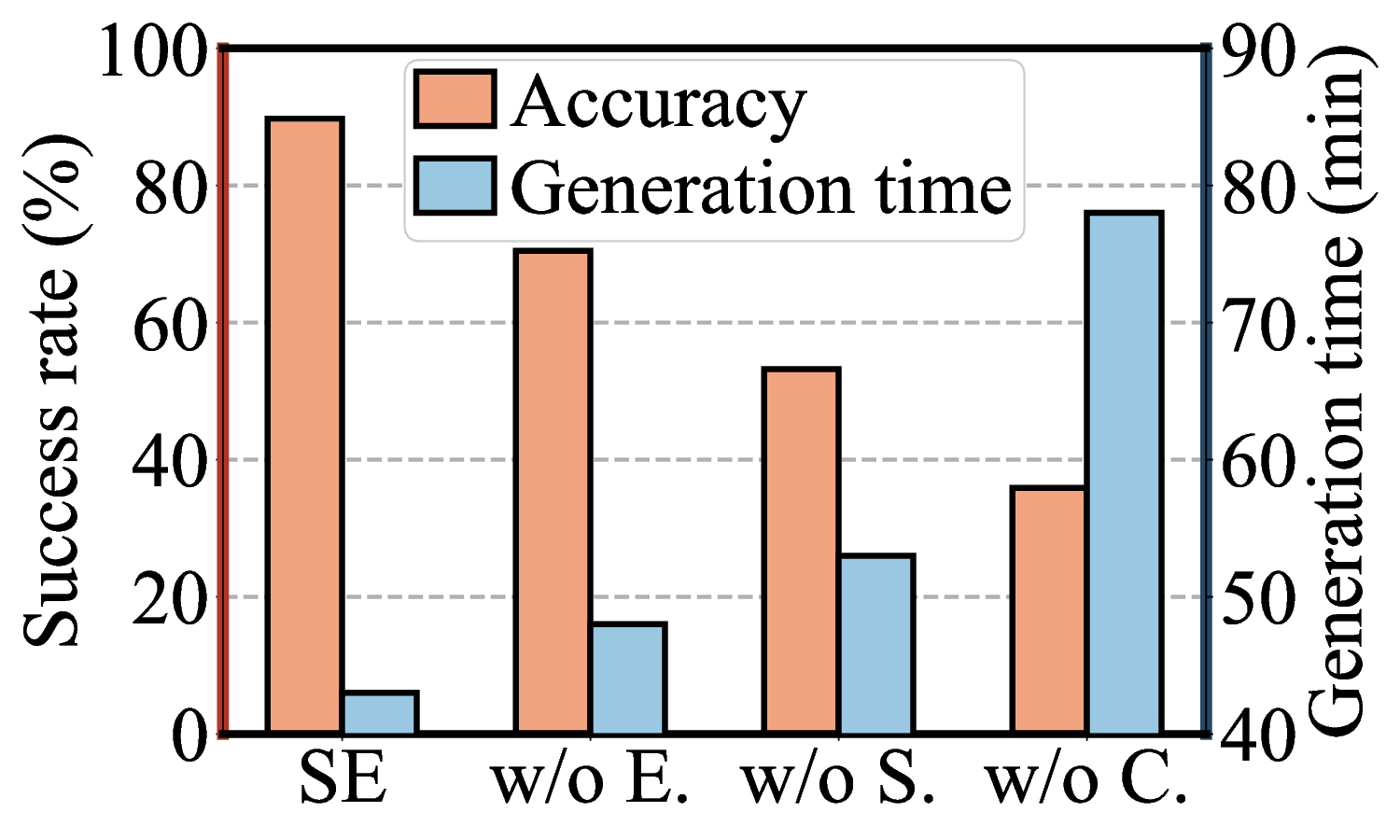}
        \label{fig:ablation_continous_llm_tuning}
    }
    \caption{Ablation study (F. -- feedback mechanism, E. -- ensemble mutual-actor-critic strategy, S. -- self-reflective grader, C. -- continuous LLM tuning).}
\end{figure}

\subsubsection{\textbf{Evaluation on VLU}}
\label{sec:AD_evaluation}
We request to deploy the VLU model on Jetson Nano. Unlike the x86-based PC where \project operates, Jetson Nano uses ARM CPUs with a distinct runtime environment for on-device deployment. Thus, agents must realize this case and adopt an appropriate inference engine to ensure compatibility with Jetson Nano.

As shown in Fig.~\ref{fig:performance_comparison_AD}, \project outperforms baselines in terms of both success rate (52.8\% $\uparrow$) and answer correct rate (23.9\% $\uparrow$). Notably, the success rate of baselines drops dramatically compared to HAR. After analyzing their intermediate outputs, we find that they all ignore runtime differences and repeatedly attempt to deploy x86-based models on ARM-based Jetson Nano, as their fixed workflows ignore runtime heterogeneity. In contrast, \project realizes this issue by proactively analyzing the runtime semantics obtained from the probe program. Then, \project creates a few more agentic steps to convert the model into a platform-agnostic ONNX format \cite{onnx} and use TensorRT \cite{tensorrt} compatible with Jetson Nano for model inference. Besides, AgentSquare needs substantial time for workflow optimization (Fig.~\ref{fig:AD_generation_time}). The accumulated context causes the agent to forget the device's memory constraints (4GB VRAM), producing models consuming over 10GB of VRAM (Fig.~\ref{fig:AD_memory_inference}).

\subsubsection{\textbf{Evaluation on ODAD}}
\label{sec:ODAD_evaluation}
We request deployment on an edge laptop with 16GB VRAM. This task is highly complex as it involves four modalities of high-dimensional data. Besides, we randomly remove some data to simulate \textit{modality missing} occurred in real-world autonomous driving scenarios. Agents ought to account for this unforeseen case and build elastic AI models capable of processing input data of varying sizes. 

As shown in Fig.~\ref{fig:performance_comparison_ODAD}, \project outperforms baselines with a significant 50\% higher success rate. After reviewing its workflows, we find that though \project initially creates a model that fails to support elastic inputs, it discovers this flaw during testing. \project then agilely adjusts its workflow and modifies the model to cope with varying input data sizes. \textit{Such developing pipelines closely resemble humans, exhibiting remarkable adaptability of \project in handling unexpected cases.} In contrast, the baselines repeatedly refine the code by randomly adopting other data processing methods without being aware of the crux. Consequently, they spend more time on debugging and produce high-memory-cost models that fail to meet resource constraints.

\subsection{Ablation Study}
\label{sec:ablation_study}
To separately test the impact of distinct modules, we conduct an ablation study on multimodal HAR, with similar phenomena being observed in other applications. Note that the superior performance of \project compared with the single-agent setting in \S~\ref{sec:feasibility_study} has demonstrated the importance of our \textit{coordinator-actor agentic architecture}.

\noindent\textbf{Mobile Runtime Feedback Mechanism.}
We test 3 ablation cases: \project, \project without the probe but replaced by a widely adopted runtime monitor psutil \cite{psutil}, and \project without any feedback. Fig.~\ref{fig:ablation_agentic_recovery_mechanism} reveals that without full feedback, \project experiences a significant 66\% drop in success rate and requires twice as long to generate executable code. Without probe-collected runtime semantics, the \textit{coordinator} cannot provide useful feedback to atomic agents, and thus cannot identify if the code is well aligned with the device's runtime environment. Though psutil can provide detailed runtime information, \project often struggles to extract content relevant to current coding subtasks, leading to a 1.5x increase in synthesis time. \textit{These results highlight the importance of our feedback mechanism in providing runtime semantics grounded in mobile devices.}

\noindent\textbf{Continuous LLM Tuning.}
We test 4 ablation cases: \project, \project without the \textit{ensemble mutual-actor-critic strategy}, \project without the \textit{self-reflective grader}, and \project without continuous tuning. In Fig.~\ref{fig:ablation_continous_llm_tuning}, all modules are crucial for retaining a high success rate and reducing code generation time. Without the mutual-actor-critic strategy, the \textit{coordinator} and atomic agents cannot be mutually updated, resulting in obsolete reasoning abilities when facing new edge AI knowledge and tasks. Without the self-reflective grader, the \textit{programmer} cannot reflect on its generated algorithms and continuously evolve to produce edge-AI-relevant code. \textit{These results show the effectiveness of our continuous LLM tuning paradigm in enhancing agents' reasoning abilities for edge AI progressively.}

\begin{table}[t]
\centering
\scriptsize
\caption{Participant information of user study (N=20)}
\label{tab:user_information}
\begin{tabular}{ll}
\toprule[1pt]
\textbf{Category}                             & \textbf{Background Information}                                                                 \\ \toprule[1pt]
\textbf{Gender}                               & Female (45\%), Male (40\%), Prefer not to say (15\%)\\ 
\textbf{Age}                                  & Under 18 (5\%), 18-30 (75\%), 30-39 (10\%), 40 and older (10\%)           \\ 
\textbf{Education}                      & Bachelor (5\%), Master (20\%), Doctoral (70\%), Others (5\%) \\ 
\textbf{English}     & Fluent (10\%), Advanced (65\%), Intermediate (20\%), Beginner (5\%)  \\ 
\bottomrule[1pt]
\end{tabular}
\end{table}

\subsection{User Study for Generalizability Test}
\label{sec:user_study}
To assess the utility and overall user satisfaction, we conduct a user study\footnote{Approved by our institution's ethics review board (IRB).} involving 10 expert users and 10 non-expert users. Expert users have rich experience in sensor data processing and AI technologies, while non-expert users only possess basic knowledge of edge AI applications. \textit{Importantly, to test the \textbf{generalizability} of \project, we did not limit the tasks. Instead, users are encouraged to freely express their requirements for developing various edge AI applications.} Table~\ref{tab:user_information} lists the detailed background information of the participants.

\subsubsection{\textbf{Metrics}}
\textit{Overall code performance} \textbf{(OCP)} evaluates the overall performance (\eg, accuracy, resource consumption) of the generated code; \textit{Code readability} \textbf{(CR)} measures the code clarity and structure; \textit{Generation efficiency} \textbf{(GE)} accesses how efficiently \project is when producing the final code; \textit{Code modularity} \textbf{(CM)} judges if the code is properly modularized for easy reuse and extension; \textit{User satisfaction} \textbf{(US)} captures users' overall experience with \project. We ask users to rate these metrics on a scale from 1 (not at all) to 5 (completely).

\subsubsection{\textbf{Overall Performance}}
Users' improvisational requests span a range of edge AI topics: mobile sensing (40\%), actuator control (40\%), and personalized services (20\%). Experts often challenge \project with more complex tasks (\eg, multimodal emotion analysis) while non-experts tend to propose simple tasks (\eg, IMU-based HAR).
In Fig.~\ref{fig:user_study_overall_performance}, both experts and non-experts provide high scores (above 4.7) for OCP, CR, CM, and US. Notably, nearly all users give CR full marks. After analyzing the generated code, we attribute it to the hierarchical agentic architecture, where distinct agents exploit their own reasoning abilities to progressively generate modular code for various functionalities. \textit{This process is akin to how humans divide a task into multiple modules and generate modular code for each.}
Besides, experts tend to rate GE lower than non-experts, as they are knowledgeable in edge AI and have higher expectations for system performance.
Despite this, GE increases as system utilization grows (\S~\ref{sec:user_study_long_term}), benefiting from the \textit{continuous tuning paradigm}.
\textit{These results exhibit the \textbf{generalizability} of \project in exploring distinct workflows for diverse edge AI tasks with enhanced user experience.}

\subsubsection{\textbf{Long-Term Evaluation}}
\label{sec:user_study_long_term}
\begin{figure}[t]
    \centering
    \subfigure[Overall performance]{
        \includegraphics[width=0.225\textwidth]{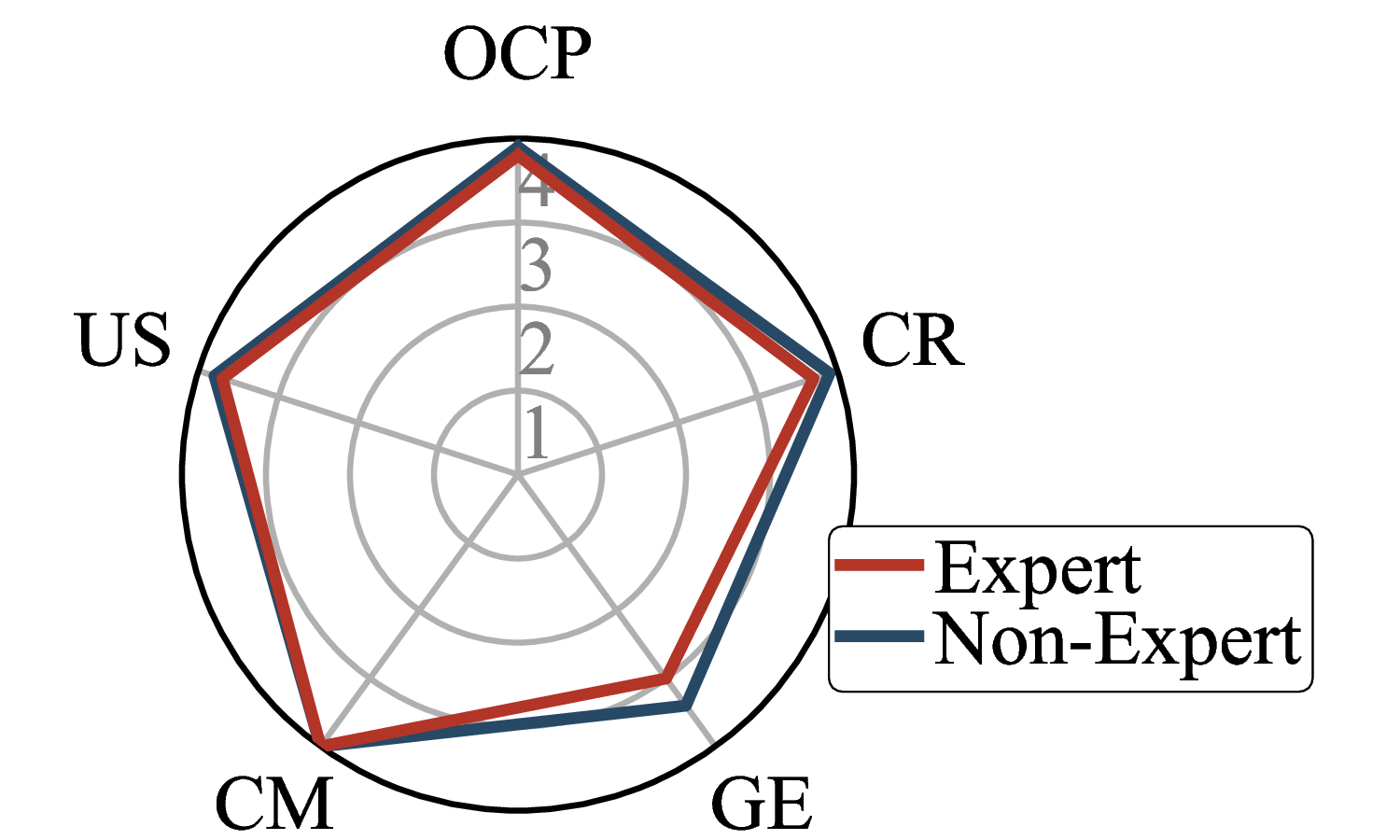}
        \label{fig:user_study_overall_performance}
    }
    \centering
    \subfigure[Long-term evaluation]{
        \includegraphics[width=0.225\textwidth]{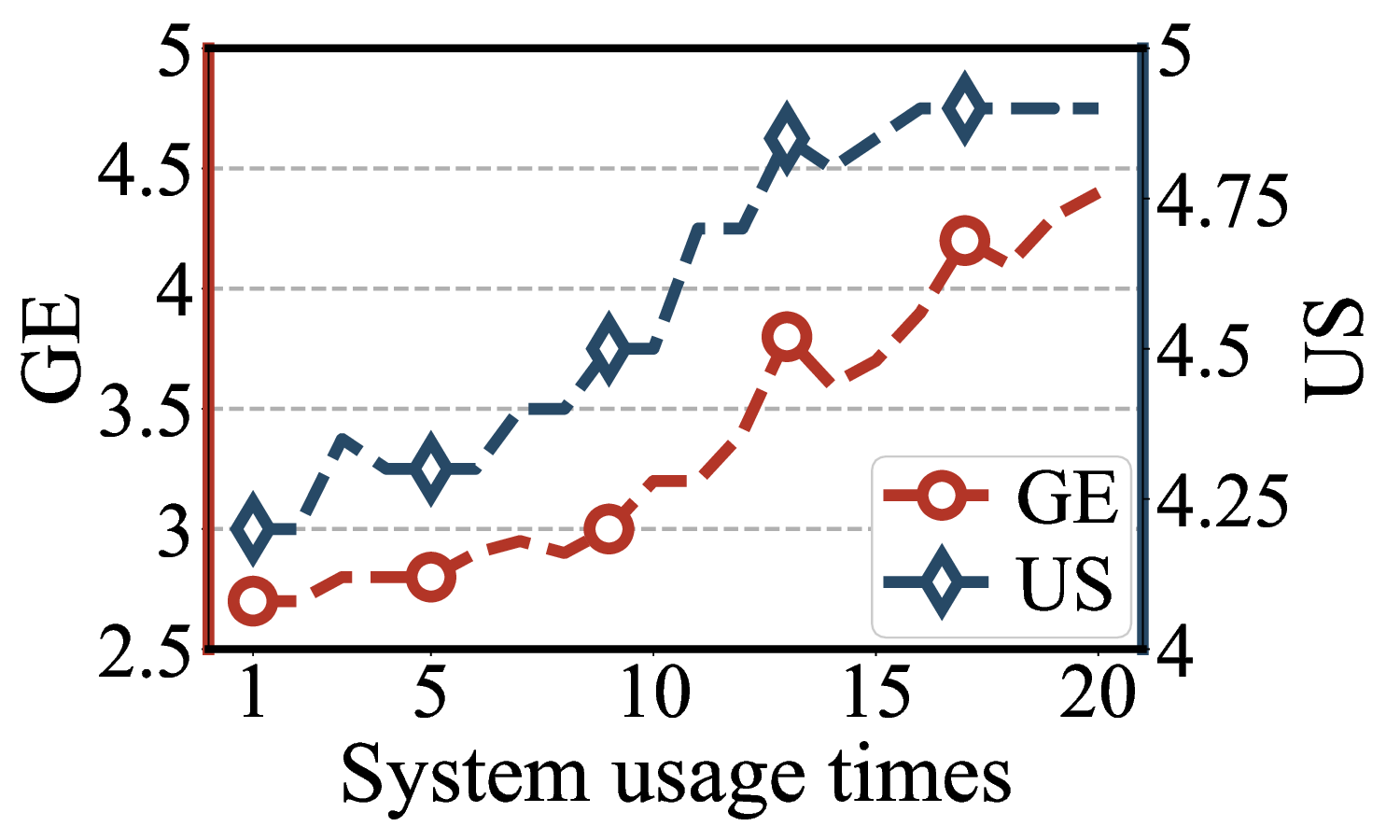}
        \label{fig:user_study_long_term_evaluation}
    }
    \caption{Evaluation results of the user study.}
\end{figure}

We further ask expert users to join in long-term evaluations to test the impact of the self-evolving paradigm. Fig.~\ref{fig:user_study_long_term_evaluation} shows that as user engagement increases over time, both GE and US rise with a similar growth trend. As system usage time increases, the scores graded by the \textit{coordinator} and atomic agents gradually accumulate and are used in mutual critique to enhance reasoning. Moreover, the \textit{retriever} progressively stores new knowledge in the local database for reference. When new similar tasks occur, \project can activate fast thinking to make decisions rapidly without re-exploration, leading to shorter generation times. Notably, though \project initially tends to adopt hallucinated algorithms, they are corrected by the feedback mechanism, and the likelihood decreases as RFT progresses. \textit{These results demonstrate the outstanding self-evolving abilities of \project in handling diverse user tasks and new edge AI knowledge.}

\section{Discussion}

\noindent\textbf{Robustness and Reliability}.
\project relies on LLMs, which may produce hallucinations during multi-step reasoning. To mitigate this issue, \project incorporates Reflexion \cite{shinn2023reflexion}, which reflects on failed attempts and evolves iteratively. The significantly higher execution success rate (\S~\ref{sec:case_study}) validates the practical reliability of \project. Moreover, to further enhance the robustness of \project in detecting and recovering from mistakes, we plan to integrate self-reflection techniques like ReAct \cite{yao2023react} as future work.

\noindent\textbf{Generalizability}.
In the user study, we invited 20 participants with distinct expertise in edge AI development. Admittedly, recruiting a large cohort of more diverse participants could more comprehensively evaluate both overall and long-term performance. In order to evaluate external validity (\ie, generalizability to diverse user requirements), we asked the participants to propose improvisational requirements. The evaluation results (\S~\ref{sec:user_study}) demonstrate the generalizability of \project in handling diverse edge AI tasks.

\noindent\textbf{Vibe Coding vs. \project}.
Vibe coding refers to using LLMs to generate code snippets purely from user instructions, without execution and verification \cite{sarkar2025vibe}. Users must manually design multiple prompts and verify the correctness of all generated code. In contrast, \project provides an end-to-end solution that automatically translates user requests into executable, runtime-compatible, and performant on-device code. By fully utilizing LLMs' reasoning abilities, \project autonomously determines and adaptively adjusts workflows for robust edge AI development.

\section{Related Work}

\noindent\textbf{Agents \& Code Generation.}
By equipping LLMs with various external tools, AI agents can automate complex tasks, such as phone manipulation \cite{lee2024mobilegpt} and IoT data interpretation \cite{xu2024penetrative}. Many works build automated agents to synthesize various IoT programs. For example, AutoIOT \cite{shen2025autoiot} generates IoT programs through IoT knowledge retrieval and iterative refinement. AutoEmbed \cite{yang2026autoembed} introduces a hardware-in-the-loop paradigm that enables agents to flash generated code onto embedded devices. However, developers must meticulously design and optimize agentic workflows to ensure that agents can follow explicitly predefined guidance, limiting their flexibility and adaptability. Instead, \project exploits LLMs' own reasoning abilities to dynamically construct workflows.

\noindent\textbf{Agentic Workflow Optimization.}
Some studies \cite{hu2025automated, zhang2025aflow} focus on automated optimization of predefined workflows to reduce human effort. For instance, ADAS \cite{hu2025automated} and AgentSquare \cite{shang2025agentsquare} regard predefined workflows as combinations of various agentic modules (\eg, Chain-of-Thought \cite{wei2022chain} and Tree-of-Thought \cite{yao2023tree}). Then, they perform an exhaustive search within constrained spaces to identify the optimal combination that maximizes performance. Yet, they still rely on fixed workflows bound by developers' experiences. Instead, drawing strengths from these arts, \project shifts the burden of workflow design from developers to LLM agents, enabling more flexible and adaptive edge AI development. Notably, \project can also be orthogonally integrated with these arts for further optimization.

\section{Conclusion}

\project is a self-exploring and self-evolving paradigm for adaptive edge AI development. By fully exploiting the reasoning abilities of LLMs, \project agents can autonomously explore distinct appropriate workflows for various edge AI tasks. In addition, our \textit{continuous LLM tuning} paradigm progressively strengthens agents to tackle evolving edge AI knowledge and environments. Extensive evaluations demonstrate the \textit{effectiveness}, \textit{generalizability}, \textit{flexibility}, and \textit{adaptability} of \project in handling diverse and evolving edge AI tasks. Looking ahead, we envision \project as a stepping stone toward more autonomous agentic systems capable of adapting to dynamic environments as well as evolving through lifelong learning and reflection.

\bibliographystyle{IEEEtran}

\clearpage


\end{document}